\documentclass{article}
\usepackage{colortbl}
\usepackage[table,xcdraw]{xcolor}
\usepackage{relsize}
\usepackage[normalem]{ulem}
\usepackage{ulem} 
\usepackage{colortbl}
\usepackage{lipsum}
\usepackage{mdframed}
\usepackage{soul}
\usepackage{PRIMEarxiv}
\usepackage{todonotes}
\usepackage{multirow}
\usepackage{diagbox} 
\usepackage{pdflscape}
\usepackage{adjustbox}
\usepackage{graphicx}

\usepackage{subcaption} 
\usepackage[utf8]{inputenc} 
\usepackage[T1]{fontenc}    
\usepackage{hyperref} 
\usepackage{url}            
\usepackage{booktabs}       
\usepackage{amsfonts}       
\usepackage{nicefrac}       
\usepackage{colortbl}  
\usepackage{longtable}
\usepackage{microtype}      
\usepackage{lipsum}
\usepackage{flushend}
\usepackage{tabularray}
\usepackage{longtable}
\usepackage{fancyhdr}       
\usepackage{graphicx}       
\graphicspath{{media/}}     
\usepackage{graphicx}
\usepackage{pdflscape}
\usepackage[compact]{titlesec}
\usepackage{rotating}
\usepackage{caption} 
\usepackage{caption}      
\usepackage{subcaption}   
\usepackage{amsmath}
\title{Leveraging Audio and Text Modalities in Mental Health: A Study of LLMs Performance}

\author{ Abdelrahman A. Ali$^\dagger$, Aya E. Fouda$^\dagger$, Radwa J. Hanafy$^{\dagger,\ddagger}$ and  Mohammed E. Fouda$^\dagger$ \\
$^\dagger$Compumacy for Artificial Intelligence solutions, Cairo, Egypt.\\
$^\ddagger$ Department of Behavioural Health- Saint Elizabeths Hospital, Washington DC, 20032.\\
fouda@compumacy.com
}

\begin{document}
\maketitle

\begin{abstract}
Mental health disorders such as depression and PTSD are increasingly prevalent worldwide, creating an urgent 
need for innovative tools to support early diagnosis and intervention. This study investigates the potential of Large Language Models (LLMs) in multimodal diagnostics for mental health disorders, specifically depression and PTSD. Using the E-DAIC dataset, we evaluate performance across text and audio modalities, as well as their integration, using zero-shot inference without task-specific fine-tuning.
The best performance for binary depression classification was achieved by Deepseek\_R1, reaching a Balanced Accuracy (BA) of 79.8\% and an F1 score of 0.75 using text input. For PTSD classification, GPT-4o mini attained a BA of 77\% and an F1 score of 0.68. Significantly, models that support multimodal input consistently benefited from combining text and audio. For example, Gemini 2.5 Pro achieved an F1 score of 0.77 and BA of 77.8\% when both modalities were integrated, representing a 3\% improvement over text-only and 2.3\% over audio-only inputs.
This paper further investigates how modality fusion impacts models differently across binary, severity, multiclass, and multi-label tasks. It also introduces new evaluation metrics—Modal Superiority Score and Disagreement Resolution Score (DRS)—to assess model reliability and complementarity. These findings demonstrate the promise of LLMs, particularly in multimodal configurations, for enhancing clinical mental health assessments.
\end{abstract}


\section{Introduction}
In recent years, mental health disorders have become increasingly prevalent, with conditions like depression and PTSD affecting a significant portion of the global population. According to the World Health Organization (WHO), over one billion people currently live with a mental disorder, with cases of depression and anxiety rising by more than 25\% during the first year of the COVID-19 pandemic. PTSD, which impacts individuals exposed to traumatic events, continues to pose significant long-term risks to well-being. In response to these escalating numbers, artificial intelligence (AI), particularly machine learning\cite{articlee}, has emerged as a vital tool, aiding in the detection and diagnosis of mental health conditions. AI models now analyze patterns in speech, behavior, and medical data, allowing for earlier intervention and improved treatment outcomes.

In recent years, models like BERT\cite{devlin2019bertpretrainingdeepbidirectional} have advanced rapidly, demonstrating significant capabilities in tasks ranging from natural language processing to decision-making in specialized domains. \cite{naveed2024comprehensiveoverviewlargelanguage} highlights the capabilities of LLMs as they are trained on vast and diverse corpora, enabling them to capture complex patterns in language. Their training allows them to generate coherent and contextually relevant text, making them highly adaptable across various applications. In healthcare, for example, LLMs can assist in diagnosing diseases, summarizing patient records, and even providing support for therapeutic interventions. Their ability to generalize across tasks has made them valuable in different fields, significantly enhancing efficiency and scalability.

LLMs show a remarkable ability to process and analyze vast amounts of text data, identifying and predicting psychiatric conditions by detecting patterns and subtle linguistic cues in patient communication. They excel at providing timely, scalable, and personalized assessments, aiding mental health professionals in diagnosis and treatment planning. LLMs provide scalable, efficient, and objective evaluations, enhancing diagnostic accuracy and personalization of treatment. However, they must be used carefully, as they may generate inaccurate responses or lead to over-reliance, requiring continuous monitoring to ensure safety and effectiveness\cite{Obradovich2024}\cite{Stade2024}.

Recent LLM advancements have enabled models to process not only text but also audio inputs directly \ cite {fang2024llamaomniseamlessspeechinteraction}, providing valuable insights into mental health. Some models convert audio to text via a speech recognition system \cite{gomez2024transformingllmscrossmodalcrosslingual}, analyzing the linguistic content for symptoms of mental disorders. However, newer models can directly analyze audio by detecting differences in tone, cadence, and speech patterns, which can reflect emotional states more accurately than text alone \cite{kong2024audioflamingonovelaudio}. This approach allows for deeper insights into a patient’s mental health, potentially leading to more precise and early diagnosis of conditions like depression or anxiety.

In this paper, we compare two approaches for mental illness detection: text modality and audio modality, using different models to analyze each. Models processing text evaluate written transcripts to identify linguistic patterns, while models processing audio analyze direct raw speech to capture features such as tone and cadence. Additionally, we explore the integration of both modalities to determine if this enhances model performance. To quantitatively assess how the combined modality affects performance, we utilize our specially formulated metrics: the MSS and DRS. By evaluating both individual and integrated modalities, we aim to provide a comprehensive overview of the potential that audio-based inputs and multimodal approaches hold for LLMs.

Also in this study, we utilized few-shot learning with the three most consistent models to evaluate their performance across text and audio modalities. Few-shot learning was applied to text-based prompts for both modalities, but inference was conducted separately for text and audio inputs. This setup allows us to compare the models’ adaptability and performance when handling text versus audio in few-shot conditions, offering insights into how each modality responds to minimal training data within the few-shot framework.

To our knowledge, there are no papers that directly input raw audio interviews into an LLM; however, some studies have constructed transformer models that can process audio inputs. Despite this, no research has yet leveraged pre-trained LLMs designed to handle audio directly for mental illness detection. Current literature lacks models capable of processing long-form audio inputs, and there has not been a comprehensive review of how small prompt changes impact LLM performance. Many studies rely on preprocessing audio data and using specific audio features for their models. In contrast, our approach aims to explore the use of raw audio inputs directly with LLMs. Furthermore, while there are LLMs that can process audio, most are restricted to segments no longer than 30 seconds. Only a few can handle slightly longer durations, but they are still not sufficient for analyzing full-length interviews effectively.

This study addresses those gaps by:

\begin{itemize}
\item performing zero-shot (ZS) preprocessing and evaluation of multiple LLMs across diverse tasks, including multi-label classification;
\item systematically comparing distinct LLM families and examining how prompt design shapes their performance;
\item analyzing results for audio and text modalities—individually and in combination—using bespoke metrics (MSS and DRS);
\item investigating how few-shot (FS) learning affects model performance across the various tasks.
\end{itemize}

This paper also examines the models’ limitations, with particular attention to how prompt wording can influence their accuracy. Subtle changes in phrasing can lead to markedly different outputs \cite{hanafi2024comprehensiveevaluationlargelanguage}, a crucial factor when evaluating LLMs. Section \ref{Methodology} details the comparison strategy for both text- and audio-based inputs, while Section \ref{Experimental Setup} outlines the experimental tasks used to gauge performance. Section \ref{Results} then presents and interprets the findings, highlighting each model’s key strengths and weaknesses.

\section{Related Work}

\begin{figure}[htb]
    \centering
    \includegraphics[width=1\linewidth]{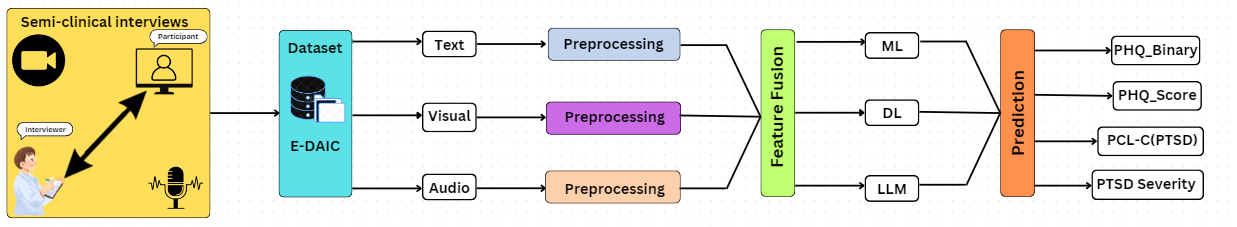}
    \caption{A visual representation of the workflow on DAIC-WOZ dataset using multimodal features}
    \label{fig:rw-fig}
\end{figure}

In most studies utilizing the DAIC-WOZ dataset\cite{Gratch2014TheDA}, researchers leverage various modalities, such as text, audio, and video, to provide a comprehensive analysis of psychological states. Each modality offers distinct insights: text data can reveal linguistic patterns, audio can capture speech characteristics, and video can provide visual cues. Depending on the study's objectives, researchers may focus on a single modality or combine multiple ones to capture a more holistic view of the participant's mental health. The following figure (Figure \ref{fig:rw-fig}) represents the workflow of most studies written on the DAIC-WOZ dataset, illustrating how these modalities are processed and integrated for prediction tasks.

To process these modalities, researchers employ techniques from machine learning (ML), deep learning (DL), and increasingly, LLMs. ML methods are often used for structured data and straightforward predictions, as detailed in Section \ref{ML}. At the same time, DL techniques, like neural networks, can handle unstructured data such as raw audio or video, as detailed in Section \ref{DL}. LLMs are particularly useful for analyzing text data, as they can understand and generate human language in a context-aware manner, making them highly effective for assessing linguistic and semantic features within mental health interviews, as detailed in Section \ref{LLM}.

Combining modalities is a common practice to enhance prediction accuracy and robustness. By integrating information from text, audio, and video, models can capture a broader spectrum of emotional and behavioral signals, leading to more reliable and nuanced assessments. This multimodal approach allows the model to compensate for the weaknesses of individual modalities and strengthen its ability to detect mental health conditions.

The goal of these approaches is typically to predict either a binary classification (e.g., whether a participant is depressed or not) or to provide a severity score for conditions like depression or PTSD. Multimodal systems combined with advanced ML, DL, and LLM techniques help deliver more accurate and comprehensive predictions, ultimately improving mental health diagnostics.

All the scores presented in Table \ref{tab:1} are reported on the DAIC-WOZ dataset, except for those that explicitly mention the E-DAIC dataset. 

\subsection{Modality Preprocessing}
Preprocessing is an essential step in working with multimodal data from the E-DAIC and DAIC-WOZ datasets. Each modality: text, audio, and visual requires different techniques to ensure the data is clean and properly formatted for model training. Preprocessing helps to remove noise, extract relevant features, and standardize the inputs across modalities. In some cases, preprocessing techniques are combined to improve the model's ability to learn relationships between different types of data, such as synchronizing audio and visual features for better context understanding.

\renewcommand{\arraystretch}{1.5} 

\begin{landscape}
\begin{table}[h]
\centering
\caption{Summary of recent models for multimodal mental illness detection, showing their years, modalities used, and reported performance metrics on datasets such as DAIC-WOZ and E-DAIC.}
\label{tab:1}
\resizebox{1.4\textwidth}{!}{%
\begin{tabular}{|c|c|c|c|c|}
\hline
Paper &
  Architecture &
  Year &
  Modalities &
  Reported Performance \\ \hline
\cite{WANG2024103106}(David Gimeno-Gómez et al.) & 
  multimodal temporal model processing non-verbal cues &
  2024 &
  Audio, Visual & \begin{tabular}[c]{@{}c@{}}
  F1 Scores: 0.67 on DAIC-WOZ,\\ 0.56 on E-DAIC dataset \end{tabular}\\ \hline
\cite{ioannides2024densityadaptiveattentionbasedspeech}(Jinhan Wang et al.) &
  Speechformer-CTC &
  2024 &
  Audio &
  F1-score of 83.15\% \\ \hline
\cite{patapati2024integratinglargelanguagemodels}(Georgios Ioannides et al.) &
  DAAMAudioCNNLSTM and DAAMAudioTransformer &
  2024 &
  Audio &
  F1 Score: 81.34\% \\ \hline
  
\cite{GUPTA2025101710}(Rohan Kumar Gupta et al.) & \begin{tabular}[c]{@{}c@{}}
  Multi-task learning (MTL) \\ with DepAudioNet and raw audio models \end{tabular} &
  2024 &
  Audio &
  \begin{tabular}[c]{@{}c@{}}F1-Score for MDD detection was 0.401 (DepAudioNet) \\  0.428 (Raw Audio) on the DAIC-WOZ dataset.\end{tabular} \\ \hline
\cite{Wu_2024}(WenWu, Chao Zhang, Philip C. Woodland) &
  Bayesian approach using a dynamic Dirichlet prior &
  2024 &
  Audio &
  F1-score: 0.600 \\ \hline
\cite{anand2024depressiondetectionanalysisusing}(Avinash Anand et al.) &
  LLMs integrating textual and audio-visual modalities &
  2024 &
  Text, Audio, Visual &
  accuracy of 71.43\% on E-DAIC \\ \hline
\cite{Huang2024}(Xiangsheng Huang et al.) &
  Wav2vec 2.0 with a fine-tuning network &
  2024 &
  Audio &
  \begin{tabular}[c]{@{}c@{}}Binary Classification Accuracy: 96.49\%\\ Multi-Classification Accuracy: 94.81\%\end{tabular} \\ \hline
\cite{Zhang2024}(Xu Zhang et al.) &
  Integration of Wav2vec 2.0, 1D-CNN, and attention pooling &
  2024 &
  Audio &
  F1-score: 79\% \\ \hline
 
\cite{burdisso2024daicwozvalidityusingtherapists}(Sergio Burdisso et al.) &
  BERT-based Longformer and Graph Convolutional Network (GCN) &
  2024 &
  Text &
  F1-score of 0.90 \\ \hline
\cite{doi:10.1080/18824889.2024.2342624}(Bakir Hadzic et al.) &
  Comparison of NLP models (BERT, GPT-3.5, GPT-4) &
  2024 &
  Text &
  \begin{tabular}[c]{@{}c@{}}BERT: F1 score: 0.59\\ GPT-3.5: F1 score: 0.78\\ GPT-4: F1 score: 0.71\end{tabular} \\ \hline
\cite{lorenzoni2024assessingmlclassificationalgorithms}(Giuliano Lorenzoni et al.) &\begin{tabular}[c]{@{}c@{}}
  Random Forest and XGBoost\\(using Sentiment Analysis and other NLP techniques) \end{tabular}&
  2024 &
  Text &
  Accuracy of 84\% \\ \hline
\cite{zhang2024llmsmeetsacousticlandmarks}(Xiangyu Zhang et al.) & \begin{tabular}[c]{@{}c@{}}
  Integration of acoustic landmarks with Large\\ Language Models (LLMs) for multimodal depression detection\end{tabular} &
  2024 &
  Audio, Text &
  F1-score of 0.84 \\ \hline
\cite{Yang2024}(Shanliang Yang et al.) &\begin{tabular}[c]{@{}c@{}}
  RLKT-MDD (Representation Learning and Knowledge \\Transfer for multimodal Depression Diagnosis)\end{tabular} &
  2024 & Text, Audio, Visual
   &
  F1 score: 80 \\ \hline
\cite{10.3389/fpsyt.2023.1160291}(Clinton Lau et al.) &
  Prefix-tuning with large language models &
  2023 &
  Text &
  \begin{tabular}[c]{@{}c@{}}(RMSE) of 4.67\\ (MAE) of 3.80\end{tabular} \\ \hline
\cite{wei2022multimodaldepressionestimationbased}(Ping-Cheng Wei et al.) &
  Sub-attentional ConvBiLSTM &
  2022 &
  Audio, Visual, Text &
  accuracy of 82.65\% and an F1-score of 0.65 \\ \hline
\cite{ghadiri2022integrationtextgraphbasedfeatures}(Nasser Ghadiri et al.) &\begin{tabular}[c]{@{}c@{}}
  Integration of text-based voice\\ classification and graph transformation of voice signals\end{tabular} &
  2022 &
  Audio, Text &
  accuracy of 86.6\% F1 of 82.4\% \\ \hline
\cite{dinkel2020textbaseddepressiondetectionsparse}(Heinrich Dinkel, Mengyue Wu, Kai Yu) &
  Multi-task BGRU network with pre-trained word embeddings &
  2020 &
  Text &
  Macro F1 score of 0.84 \\ \hline
\cite{xezonaki2020affective}(Danai Xezonaki et al.) &
  Hierarchical Attention Network with affective conditioning &
  2020 &
  Text &
  68.6 F1 scores \\ \hline
\cite{stepanov2017depressionseverityestimationmultiple}(Evgeny Stepanov et al.) & \begin{tabular}[c]{@{}c@{}}
  multimodal system utilizing speech, \\ language, and visual features \end{tabular}&
  2017 &
  Audio, Text, Visual & \begin{tabular}[c]{@{}c@{}}
  PHQ-8 results with\\ a Mean Absolute Error (MAE) of 4.11 \end{tabular}\\ \hline
\end{tabular}%
}
\end{table}
\end{landscape}

\subsubsection{Text Preprocessing} In working with textual data from multimodal datasets, several preprocessing techniques are commonly employed to ensure the text is clean and structured before further analysis. Below are the primary preprocessing techniques applied to textual data, along with the papers that have utilized these techniques:
    
    \begin{itemize}
        \item \textbf{Basic Text Cleaning:} This includes the removal of irrelevant annotations, such as speaker tags, hardware syncing notes, and non-verbal cues (e.g., laughter). Text is often lowercased to ensure uniformity, and punctuation is standardized to maintain semantic context. Papers such as those by \textit{Rohan Kumar Gupta et al. (2024)}\cite{GUPTA2025101710} and \textit{Ping-Cheng Wei et al. (2022)}\cite{wei2022multimodaldepressionestimationbased} have implemented this step, ensuring that the cleaned text is ready for further processing.

        \item \textbf{Tokenization and Removal of Stop Words:} Tokenization is a crucial step in breaking down transcriptions into words or smaller linguistic units. Stop words (common words such as "the" or "and") are often removed to focus on more meaningful terms in the dataset. Several papers, including those by \textit{Clinton Lau et al. (2023)}\cite{10.3389/fpsyt.2023.1160291} and \textit{Giuliano Lorenzoni et al. (2024)}\cite{lorenzoni2024assessingmlclassificationalgorithms}, applied tokenization and stop word removal to improve model performance by focusing on more significant features within the text.

        \item \textbf{Feature Extraction using Embeddings:} After cleaning and tokenization, the textual data is often transformed into feature vectors using pre-trained language models or embeddings such as BERT, GloVe, or Word2Vec. This allows for capturing the deeper semantic meaning of the text. The papers by \textit{Avinash Anand et al. (2024)}\cite{anand2024depressiondetectionanalysisusing} and \textit{Bakir Hadzic et al. (2024)}\cite{doi:10.1080/18824889.2024.2342624} employed BERT embeddings, while other papers, such as \textit{Xiangyu Zhang et al. (2024)}\cite{zhang2024llmsmeetsacousticlandmarks}, utilized GloVe and Word2Vec embeddings to capture contextual and lexical features from the transcriptions.
    \end{itemize}

    These preprocessing steps are essential to ensuring that the textual data is in a format suitable for downstream machine learning models, allowing them to accurately detect and predict mental health conditions based on language patterns. 
   \subsubsection{Audio Preprocessing} Audio data in multimodal datasets undergoes various preprocessing steps to ensure high-quality inputs for different models. These steps include cleaning the audio, extracting meaningful features, and normalizing the data. Below are the primary preprocessing techniques applied to audio data, along with the papers that have utilized these techniques:
    
    \begin{itemize}
        \item \textbf{Resampling and Noise Removal:} To standardize the audio data, many studies resample it to a consistent frequency, often 16 kHz, and remove noise, including long pauses and irrelevant sounds. For example, \textit{David Gimeno-Gómez et al. (2024)}\cite{WANG2024103106} resampled audio data and used feature extraction tools to focus on relevant sound signals. Similarly, \textit{Xiangsheng Huang et al. (2024)}\cite{Huang2024} used noise removal techniques to clean the audio before processing.

        \item \textbf{Feature Extraction with MFCCs and Spectrograms:} Mel-frequency cepstral coefficients (MFCCs) and log-mel spectrograms are commonly extracted from the audio data to capture speech and acoustic features. Papers like \textit{Jinhan Wang et al. (2024)}\cite{ioannides2024densityadaptiveattentionbasedspeech} and \textit{Rohan Kumar Gupta et al. (2024)}\cite{GUPTA2025101710} utilized MFCCs to capture essential features from the raw audio signals, while \textit{Xiangsheng Huang et al. (2024)}\cite{Huang2024} extracted log-mel spectrograms for further analysis.

        \item \textbf{Advanced Feature Extraction using pre-trained Models:} In some studies, pre-trained models such as HuBERT and wav2vec 2.0 are used to extract higher-level audio features. For instance, \textit{Avinash Anand et al. (2024)}\cite{anand2024depressiondetectionanalysisusing} used HuBERT-large to extract 1024-dimensional features from the audio, while \textit{Xu Zhang et al. (2024)}\cite{Zhang2024} applied wav2vec 2.0 for frame-level feature extraction, enhancing the model's ability to analyze complex audio patterns.
    \end{itemize}

    These preprocessing steps are crucial for transforming raw audio data into meaningful inputs, ensuring that different models can effectively analyze speech patterns, acoustic features, and other relevant audio signals for detecting mental health conditions.
    \subsubsection{Visual Preprocessing} Visual data, particularly facial expressions and body language, plays a significant role in multimodal datasets. Various preprocessing steps are applied to extract meaningful features from visual data, such as facial landmarks and action units, which are then used for mental health predictions. Below are the primary preprocessing techniques applied to visual data, along with the papers that have utilized these techniques:
    
    \begin{itemize}
        \item \textbf{Facial Landmark Detection and Normalization:} Facial landmarks, including key points such as eye, nose, and mouth positions, are extracted to understand emotional expressions. Normalization techniques are often used to ensure uniformity across different participants. For example, \textit{Ping-Cheng Wei et al. (2022)}\cite{wei2022multimodaldepressionestimationbased} extracted and normalized facial landmarks for consistency in facial expressions, and \textit{Xiangsheng Huang et al. (2024)}\cite{Huang2024} applied similar techniques to capture important facial features.

        \item \textbf{Facial Action Units (FAUs) Extraction:} Facial Action Units (FAUs) capture muscle movements that reflect various emotions, making them essential for predicting mental states. The OpenFace toolkit is commonly used to extract FAUs. Studies such as \textit{Avinash Anand et al. (2024)}\cite{anand2024depressiondetectionanalysisusing} and \textit{Rohan Kumar Gupta et al. (2024)}\cite{GUPTA2025101710} used FAUs as a key visual feature for their models, focusing on facial expressions linked to emotional and mental health states.

        \item \textbf{Pose and Head Movement Features:} In addition to facial features, head pose and body movement features are extracted to analyze non-verbal behavior. These features help to capture body language and gaze direction. \textit{Giuliano Lorenzoni et al. (2024)}\cite{lorenzoni2024assessingmlclassificationalgorithms} and \textit{Clinton Lau et al. (2023)}\cite{10.3389/fpsyt.2023.1160291} both employed techniques to capture head pose and movement features, improving their models' ability to interpret visual cues related to mental health.
    \end{itemize}

    These preprocessing techniques are essential for extracting rich, meaningful features from visual data, which are then used to predict mental health conditions by analyzing facial expressions, body language, and other non-verbal cues.

\subsection{Processing  Techniques}

Once the textual, audio, and visual data are preprocessed, different Processing techniques are applied to predict mental health conditions like depression and PTSD. These models aim to leverage the cleaned and extracted features from each modality, learning patterns that can indicate the presence of psychological distress. The most commonly used approaches include Machine Learning (ML), Deep Learning (DL), and, more recently, LLMs, each of which contributes uniquely to the field.

    \subsubsection{Machine Learning (ML)} \label{ML} Machine learning techniques primarily focus on structured feature extraction from preprocessed data. Models such as random forests, support vector machines (SVMs), and XGBoost are often employed to analyze features from text, audio, and visual modalities. For instance, \textit{Giuliano Lorenzoni et al. (2024)}\cite{lorenzoni2024assessingmlclassificationalgorithms} used Random Forest and XGBoost models to process text features like sentiment analysis and word frequency, achieving high accuracy in detecting mental illness. Similarly, \textit{Shanliang Yang et al. (2024)}\cite{Yang2024} implemented multi-task learning and knowledge transfer techniques (RLKT-MDD) to improve their multimodal depression diagnosis system. \textit{Xiangyu Zhang et al. (2024)}\cite{doi:10.1080/18824889.2024.2342624} employed machine learning techniques, specifically focusing on the integration of acoustic landmarks with language models to enhance their mental health predictions. These machine learning models are highly interpretable and work well with small to medium-sized datasets, leveraging relationships between structured features to predict mental health outcomes like depression and PTSD.

    \subsubsection{Deep Learning (DL)}\label{DL} Deep learning models, especially convolutional neural networks (CNNs), recurrent neural networks (RNNs), and other advanced architectures, are widely used for processing unstructured data, such as raw audio and visual inputs. These models are known for their ability to extract complex hierarchical patterns from the data. For example, \textit{Xiangsheng Huang et al. (2024)}\cite{Huang2024} applied a CNN-based architecture to analyze log-mel spectrograms from audio data, achieving excellent accuracy in binary classification for depression. Similarly, \textit{Xu Zhang et al. (2024)}\cite{Zhang2024} integrated Wav2Vec 2.0 with CNNs and attention pooling to fuse audio and visual modalities, demonstrating the power of DL in handling multimodal data.

    Other studies, such as \textit{Rohan Kumar Gupta et al. (2024)}\cite{GUPTA2025101710}, utilized LSTM networks to process sequential audio data, capturing temporal patterns that are indicative of depression. LSTM and RNN models are particularly effective for analyzing speech data over time, allowing for the detection of subtle emotional cues across audio sequences.

    Additionally, \textit{David Gimeno-Gómez et al. (2024)}\cite{WANG2024103106} focused on a multimodal temporal model that processes non-verbal cues from various modalities, utilizing deep learning architectures to improve predictions in mental health detection. The combination of multiple inputs, such as audio, visual, and text, through deep learning models allows for more comprehensive analyses of participant behaviors.

    \subsubsection{Large Language Models} \label{LLM}LLMs like BERT, GPT-3.5, and GPT-4 have become essential tools in analyzing text data, particularly when working with transcriptions from clinical interviews. LLMs excel at understanding the deeper semantic context and patterns within language, making them highly effective for predicting mental health conditions from text-based data. 

    For instance, \textit{Avinash Anand et al. (2024)}\cite{anand2024depressiondetectionanalysisusing} integrated LLMs with multimodal data, including textual and audio-visual modalities, to achieve better contextual understanding of patient responses. The use of BERT-based embeddings in this study enhanced the model’s ability to extract meaning from text and fuse it with non-verbal cues like facial expressions and vocal tones.

    \textit{Clinton Lau et al. (2023)}\cite{10.3389/fpsyt.2023.1160291} applied prefix-tuning to large language models, such as GPT-4, to fine-tune their performance for specific tasks like depression severity estimation. By leveraging the contextual power of LLMs, they were able to capture subtle emotional cues from the patient transcripts that traditional models might miss. 

    \textit{Bakir Hadzic et al. (2024)}\cite{doi:10.1080/18824889.2024.2342624} performed a comparison of several NLP models (BERT, GPT-3.5, GPT-4) in predicting mental health conditions, finding that transformer-based models can capture linguistic nuances in patient interviews with high precision.

    These studies demonstrate the unique ability of LLMs to handle large and complex text sequences, improving the overall accuracy of predicting mental health conditions through text analysis, especially when combined with other data modalities.

\section{Datasets}
\label{dataset}
The Extended Distress Analysis Interview Corpus (E-DAIC) is an enhanced version of the DAIC-WOZ, designed to study psychological conditions such as anxiety, depression, and PTSD through semi-clinical interviews. The interviews are conducted by a human-controlled virtual agent named "Ellie" in a wizard-of-Oz (WoZ) setting or by an autonomous AI agent, both aiming to detect verbal and nonverbal indicators of mental illnesses. Developed as part of the DARPA Detection and Computational Analysis of Psychological Signals (DCAPS) program, this dataset is specifically crafted to advance the understanding and detection of psychological stress signals, with a particular focus on depression. The dataset is available through the University of Southern California's Institute for Creative Technologies (USC ICT). It can be accessed by researchers through a data use agreement, ensuring ethical compliance and protection of participant data. Approval for the use of this dataset was obtained from USC ICT, emphasizing its adherence to institutional guidelines for studying psychological health conditions.

The dataset contains 275 samples, which are systematically divided into training, development, and test sets. This division ensures a balanced representation of participants in terms of age, gender, and depression severity, as measured by the PHQ-8 scores, with the test set consisting exclusively of sessions conducted by the AI-controlled agent. This unique structure provides an invaluable resource for evaluating autonomous interaction models in the context of mental health diagnostics.

Each session directory is structured to include various files:
\begin{itemize}
    \item  Audio recordings (WAV format)
    \item  Transcripts (CSV format)
    \item  Feature sets derived from audio and visual data:
    \item  Audio features like eGeMAPS and MFCCs processed into a bag-of-words model.
    \item  Visual features including Pose, Gaze, and Action Units (AUs), summarized over set intervals.
    \item  Deep representations from CNN models like ResNet, VGG, and Densenet for spectral images converted from audio.

\end{itemize}

The E-DAIC dataset's comprehensive structure supports Multiple viewpoints on depression by offering extensive behavioral, acoustic, and visual cues. This rich combination of data modalities allows researchers to develop and test diagnostic models that can autonomously assess psychological distress with greater accuracy. For instance, the deep learning models trained on this dataset can leverage the diverse and detailed features to identify subtle indicators of depression, enhancing the potential for early and more reliable detection of mental health issues. This approach is particularly valuable in clinical simulations and real-world applications, where automated systems can provide consistent and unbiased assessments.

\subsection{Data Analysis}

The E-DAIC dataset includes four distinct labels used to classify mental health conditions, focusing on both depression and Post-Traumatic Stress Disorder (PTSD). These labels provide a comprehensive analysis by offering both binary classification and severity scores for each condition.

\begin{itemize}
    \item \textbf{PHQ\_Binary:} This label classifies individuals based on depression using the PHQ-8, a standard questionnaire for assessing depression. In this binary classification, individuals are labeled as "Negative" or "Positive" for depression. A score of 10 or higher on the PHQ\_Score corresponds to the "Positive" label, indicating the presence of depressive symptoms.
    
    \item \textbf{PHQ\_Score:} This is a continuous score derived from the PHQ-8, ranging from 0 to 24, and it represents the severity of depression. Individuals with a score of 10 or higher are considered to have clinically significant depression. The PHQ\_Binary classification is directly based on this score, with a cutoff point at 10.
    
    \item \textbf{PCL-C (PTSD):} This binary label indicates whether an individual meets the criteria for PTSD based on the PCL-C (Post-Traumatic Stress Disorder Checklist – Civilian Version). Similar to the PHQ\_Binary, individuals are classified as "Negative" or "Positive" based on their PTSD severity score.
     
    \item \textbf{PTSD Severity:} This is a continuous score that assesses the severity of PTSD symptoms. A score higher than 44 indicates the presence of PTSD. The binary PCL-C classification is derived from this severity score, with 44 serving as the threshold for diagnosis.
\end{itemize}
Table \ref{tab:data-bin} below summarizes the count of individuals classified as "Negative" or "Positive" for both depression and PTSD, providing a binary overview of these conditions within the dataset:
\begin{table}[htb]
\centering
\caption{Distribution of Binary Classification for Depression and PTSD}
\label{tab:data-bin}
\begin{tabular}{|c|c|c|}
\hline
\textbf{Disorder} & \textbf{Negative} & \textbf{Positive} \\
\hline
PHQ Binary & 189 & 86 \\
\hline
PCL-C (PTSD) & 188 & 87 \\
\hline
\end{tabular}
\end{table}
\subsubsection{Data Preprocessing}
\label{sec:data_preprocessing}

\begin{itemize}
    \item \textbf{Label Correction:} In the E-DAIC dataset, an issue with incorrect labeling was identified in the PHQ\_Binary classification. Specifically, there are 20 instances where the PHQ\_Score is 10 or higher, indicating that the participants should be classified as "Positive" for depression. However, the PHQ\_Binary label was incorrectly assigned as 0 (Negative) instead of 1 (Positive). \textbf{The IDs of the incorrect samples ID = [320, 325, 335, 344, 352, 356, 380, 386, 409, 413, 418, 422, 433, 459, 483, 633, 682, 691, 696, 709]}. This mislabeling can lead to inaccuracies in model training and prediction if not corrected during the data preprocessing stage.

\item \textbf{Severity Mapping:} 

\begin{itemize}
\item\textbf{Depression}

Based on the PHQ-8 depression scale explained in the referenced paper \cite{articles}[Kroenke et al.], we derived the severity mapping for depression scores ranging from 0 to 24. 
    However, the labels associated with these categories were not explicitly provided in the referenced paper. We applied standard clinical terminology to label the ranges as seen in the figure.

\vspace{0.2cm}
Table \ref{tab:severity-labels} summarizes the count of participants falling within each severity label. The labels are mapped as follows: 0 refers to a PHQ\_Score between 0-4, 1 refers to scores from 5-9, and so on.

\begin{table}[htb]
\centering
\caption{Count of Participants by PHQ-8 Severity Labels}
\label{tab:severity-labels}
\begin{tabular}{|c|c|c|}
\hline
\textbf{Intervals} & \textbf{Label}       & \textbf{Count of Participants} \\ \hline
0-4: Minimal & 0           & 122             \\ \hline
5-9: Mild & 1              & 67              \\ \hline
10-14: Moderate & 2         & 43              \\ \hline
15-19: Moderately Severe & 3  & 33              \\ \hline
20-24: Severe & 4           & 10              \\ \hline
 \textbf{Total} &  & \textbf{275}        \\ \hline
\end{tabular}
\end{table}

\item \textbf{PTSD}

Based on the PCL-C PTSD scale explained in the referenced paper García-Valdez et al. (2024) \cite{10.1007/978-3-031-46933-6_21}, we derived the severity mapping for PTSD symptoms. According to the paper, the labels are used to categorize PTSD severity as follows: \begin{itemize} \item 0: little to no severity \item 1: Moderate severity \item 2: High severity \end{itemize} 
\vspace{0.2mm}
The PCL-C score intervals are chosen based on the understanding of the used LLM and are detailed in figure \ref{tab:PTSD-MAP}. The Results \& Discussion (Section \ref{Results}) discuss the scoring system and compare it to existing intervals from the literature.

\end{itemize}
\end{itemize}

\section{Methodology}
\label{Methodology}
In this section, we discuss the proposed methodology including processing pipelines, prompt engineering and  LLMs under evaluation.

\subsection{Evaluation Pipeline for Audio-Based Data}

\begin{figure}[h]

    \centering \includegraphics[width=1\linewidth]{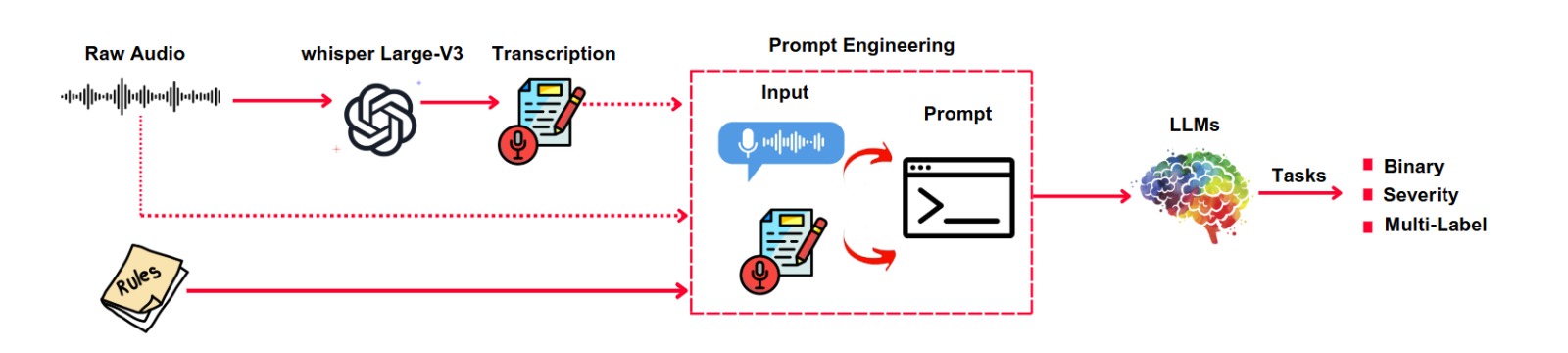}
    \caption{Proposed evaluation pipeline for audio-based data.}
    \label{fig:pipeline}
\end{figure}

The proposed evaluation pipeline for audio-based data, as illustrated in Figure~\ref{fig:pipeline}, begins with raw audio inputs. These audio samples can be directly provided to the model or first transcribed into text using the Whisper Large-V3 model. In addition, the pipeline supports integrating both modalities—raw audio and transcribed text. After determining the preferred input format (audio only, text only, or a combination of both), a prompt engineering step is conducted. Here, carefully crafted task-specific prompts guide the model toward binary classification, severity classification, or multi-label classification tasks. These prompts are designed to ensure that the model receives clear instructions, aligned with the chosen input modality or modalities.

Once the input (audio, transcription, or both) is combined with the tailored prompts, the resulting prompt is passed to LLMs for evaluation. This approach enables the assessment of the model’s zero-shot capabilities, evaluating how well it can perform classification tasks without prior fine-tuning or preprocessing. By examining LLM responses across different modalities and functions, this pipeline provides insights into the model’s inherent ability to generalize, adapt, and accurately interpret a variety of input formats and instructions.

In this setup, all 275 samples from the E-DAIC dataset are used in their entirety as a test set, ensuring a comprehensive evaluation of model performance. By comparing models across modalities—raw audio, transcribed text, and combined inputs—the evaluation highlights which modality performs better under specific conditions and tasks. This methodology helps identify optimal configurations and provides valuable insights into how the models adapt to varying input formats and task requirements.

\noindent

\subsection{Experimental Setup}
\label{Experimental Setup}
In this study, a comprehensive experimental setup was established to evaluate the effectiveness of LLMs in predicting mental health conditions, specifically depression and PTSD, using both text and audio modalities. The main objective of this experiment is to compare the performance of LLMs when processing textual data, such as transcriptions from interviews, against their performance in analyzing audio data that includes vocal features. By leveraging multimodal data, the aim is to assess how well each modality contributes to the accurate prediction of mental health conditions and whether combining them can enhance the overall predictive power of the models. In this setup, the audio modality provides information about vocal features, such as tone, pitch, and speech rate, which are indicative of emotional states. On the other hand, the text modality offers insights into the linguistic patterns and cognitive expressions of the participants, as derived from the transcriptions. Both data types are processed using state-of-the-art LLMs. Additionally, we explore the integration of both modalities to determine if this approach enhances model performance, providing a more comprehensive understanding of each participant’s mental state. This comparative study aims to provide insights into the strengths and limitations of each modality in mental health prediction and evaluate the benefits of combining them for more robust and accurate classification.

\subsubsection{Task 1: Binary Classification}

The first task in the experimental setup involves binary classification, where participants are categorized into two groups: depressed or not depressed, and PTSD-positive or PTSD-negative. For depression, the PHQ-8 (Patient Health Questionnaire-8) scores are used as a reference, with participants classified as depressed if their score meets or exceeds a predefined threshold of 10. Similarly, for PTSD, the PCL-C (Post-Traumatic Stress Disorder Checklist) scores are used, with a score threshold of 44 indicating PTSD positivity.

\subsubsection{Task 2: Severity Classification}

In this task, the focus was on classifying the severity of depression and PTSD, by categorizing the severity of depression into multiple levels based on the PHQ-8 score. The severity levels range from minimal (0-4), mild (5-9), moderate (10-14), moderately severe (15-19), to severe (20-24). Additionally, PTSD severity is classified into three categories: little or no severity, moderate severity, and high severity. This granularity enables a more detailed understanding of an individual’s mental health, facilitating tailored treatment plans corresponding to each severity level.


\subsubsection{Task 3: Multiclass Classification}
\label{task3}
In this task, we extend the classification approach by combining the binary classifications of depression and PTSD to create a multiclass framework. Participants are categorized into one of several classes based on their mental health status: no disorder, depression only, PTSD only, or both depression and PTSD. This multiclass setup allows us to evaluate the performance of Large Language Models (LLMs) in predicting whether a participant has one or more mental health disorders or none at all.

To assess the effectiveness of LLMs, we compare both the text and audio modalities, as well as their combination. This comprehensive comparison aims to identify how well each modality and their combination perform in simultaneously classifying multiple disorders, providing insights into the potential benefits of a multimodal approach in mental health diagnostics.

\subsection{Audio Handling}

For the audio handling process, we directly utilized the raw audio files from the E-DAIC dataset without applying any preprocessing or cleaning techniques. The average interview duration in the dataset is approximately 16 minutes. These unaltered audio files were essential inputs for both the analysis and transcription stages. This raw data was then used during the transcription process, ensuring that all acoustic nuances were preserved and processed by the Whisper model, as described in the transcription process section. By working with the original files, we aimed to evaluate the models in a real-world scenario, where audio imperfections such as background noise and variability in speech could impact model performance.
\subsection{Transcription Process}
For the transcription process, we used the Whisper \cite{radford2022robustspeechrecognitionlargescale} model, specifically the Large-V3 version, to transcribe the entire interview data, including both the interviewer’s prompts and the participant’s responses. The transcription provided in the dataset contained only the answers given by the participants, omitting the interviewer’s questions. By transcribing the whole interaction, we were able to capture crucial contextual information from the interviewer’s prompts (e.g., Ellie’s prompts), which can provide significant insights. These prompts often contain information that the models can exploit to classify the participants more effectively, as highlighted in previous research by Sergio Burdisso et al. (2024).

One of the key features of Whisper’s architecture is its ability to handle multilingual transcription, background noise, and various accents with high precision. Whisper processes audio data in chunks, typically by converting the audio into spectrograms (a visual representation of sound) that the neural network can then interpret. This process allows Whisper to extract meaningful patterns from the audio signal, even in challenging acoustic conditions, such as overlapping speech or background noise.

By utilizing Whisper to transcribe the full interview, we ensured that both the content and style of speech were accurately captured. This comprehensive transcription process was crucial for later stages of analysis, providing the model with richer data that could enhance classification performance. Whisper’s robust handling of varied audio conditions ensured the accuracy and reliability of the transcribed data, forming a solid foundation for the text-based models applied in subsequent analysis.
\begin{figure}[htb]
\centering
\begin{minipage}{0.9\textwidth}
\textit{
"Have you ever served in the military? No.
Have you ever been diagnosed with PTSD? Yes, I have.
How long ago were you diagnosed? In, um, I don't know. I was in the military. I was in the military.
How long ago were you diagnosed? In February of 2011.
What got you to seek help? A stalker attacked me and almost killed me in November of 2009.
He broke into my apartment and lay in wait for me, and attacked me when I came in the door, and tried to kill me.
Do you still go to therapy now? I do."
}
\end{minipage}
\caption{Part of a transcript for a sample interview}
\label{fig:sample-transcript}
\end{figure}

Figure \ref{fig:sample-transcript} represents a snippet of how both questions and answers were captured during the transcription process, allowing the models to have access to more complete data for analysis, which includes not just the participant's responses but also the context provided by the interviewer's prompts.

\subsection{LLMs Under Evaluation}
\begin{table}[h]
\centering
\caption{Summary of Models evaluated in this analysis}
\label{tab:m1}
\begin{tabular}{|l|l|l|l|l|}
\hline
Model                                                         & Parameters  & Source     & API Provider Used   & Modality      \\ \hline
Llama 3.1 70B \cite{dubey2024llama}         & 70B         & Meta       & Groq                & Text          \\ \hline
Gemma 2 9B \cite{team2024gemma2}        & 9B          & Google     & Nvidia NIM          & Text          \\ \hline
Mistral NeMo  \cite{mistral2024nemo}        & 12B         & Mistral AI & Mistral AI API      & Text          \\ \hline
GPT-4o mini \cite{openai2024gpt4omini}     & Proprietary & OpenAI     & OpenAI API          & Text          \\ \hline
Phi-3.5-mini \cite{microsoft2024phi3.5slms} & 3.8B        & Microsoft  & Nvidia NIM          & Text          \\ \hline
Phi-3.5-MoE \cite{microsoft2024phi3.5slms}  & 42B         & Microsoft  & Azure AI API        & Text          \\ \hline
Deepseek\_R1 \cite{deepseekai2025deepseekr1incentivizingreasoningcapability}& {Proprietary}  & {Deepseek}  & {Deepseek API}  &{Text}  \\ \hline
{Llama 3.3 70B \cite{dubey2024llama}  }                           & {70B}         & {Groq}      & {Groq API}            & {Text}         \\ \hline
Gemini 1.5 Flash \cite{google2024geminiapi} & 8B          & Google     & Google's Gemini API & Text \& Audio \\ \hline
Gemini 1.5 Pro \cite{google2024geminiapi} & Proprietary & Google     & Google's Gemini API & Text \& Audio  \\ \hline

{Gemini 2.5 Pro \cite{google2024geminiapi}}  & {Proprietary}  & {Google}  & {Google's Gemini API}  & {Text \& Audio} \\ \hline
{Gemini 2 Flash \cite{google2024geminiapi}}  & {Proprietary}  & {Google}  & {Google's Gemini API}  & {Text \& Audio} \\ \hline
{Gemini Flash Thinking \cite{google2024geminiapi}}  & {Proprietary}  & {Google}  & {Google's Gemini API}  & {Text \& Audio} \\ \hline
{Gemini 2 Flash Lite \cite{google2024geminiapi}}  & {Proprietary}  & {Google}  & {Google's Gemini API}  & {Text \& Audio} \\ \hline

\end{tabular}
\end{table}

In this study, we evaluated a diverse set of large language models (LLMs) to analyze their capabilities in handling both text and audio data. The models were selected based on factors such as parameter size, source, accessibility via APIs, and support for different data modalities. Additionally, models were excluded based on findings and recommendations from the referenced paper \cite{hanafi2024comprehensiveevaluationlargelanguage}. These criteria ensured a comprehensive comparison of models from various providers, including both proprietary and open-source options.

Table \ref{tab:m1} Table \ref{tab:m1} summarizes the characteristics of the selected models, which vary significantly in terms of parameter size, from lightweight models like Phi-3.5-mini (3.8B parameters) to larger ones such as Llama 3.1 and 3.3 (both 70B). The sources include major companies like Google, Microsoft, and Meta, as well as specialized firms like Mistral AI and Deepseek. API accessibility is provided through a range of platforms, including Nvidia NIM, OpenAI API, Groq, Azure AI API, and Google’s Gemini API, allowing flexible deployment options across text and audio modalities.

For the audio analysis, Whisper was employed to transcribe audio files into text before inputting the results into text-focused models like Llama 3 and GPT-4o mini. In contrast, models supporting multimodal data, such as Gemini 1.5 Flash, Gemini 1.5 Pro, and Gemini 2.5 Pro, were directly fed audio data to evaluate their performance in handling both audio and text tasks. This approach allowed for a direct comparison of text-only versus multimodal model capabilities.

Several models in this evaluation, such as Phi-3.5-mini and Phi-3.5-MoE, were chosen due to their emerging relevance in multimodal and multilingual tasks. The inclusion of models with different quantization strategies, such as those used in Llama 3.1 70B and Mistral NeMo, highlights the trade-offs between model complexity and computational efficiency. Quantization, in some cases, helped optimize model performance, particularly in audio-capable models.

\begin{table}[h]
\centering
\caption{Excluded models and associated technical issues}
\label{tab:m2}
\begin{tabular}{|l|l|l|l|l|}
\hline
Model                                                        & Parameters & Source            & API Provider Used & Modality   \\ \hline
Qwen/Qwen2 \cite{chu2024qwen2audiotechnicalreport}.          & 7B         & Qwen              & Huggingface       & Audio \& Text \\ \hline
Flamingo \cite{kong2024audioflamingonovelaudio}.             & 9B         & NVIDIA            & Huggingface       & Audio \& Text \\ \hline
Llama3-S \cite{aibase2024}                                   & 8B         & Homebrew Research & Huggingface       & Audio      \\ \hline
Llama Omni \cite{fang2024llamaomniseamlessspeechinteraction} & N/A        & N/A               & Huggingface       & Audio      \\ \hline
Mini Omni \cite{xie2024miniomnilanguagemodelshear}           & N/A        & OpenAI            & Huggingface       & Audio      \\ \hline
\end{tabular}
\end{table}

Table \ref{tab:m2} presents a summary of models that were excluded from the study because they did not meet the specific requirements needed for the analysis. The exclusion criteria were based on technical limitations that hindered the models' ability to process the dataset effectively or misalignments between the models' primary functionalities and the study's objectives. Each model had distinct reasons for exclusion, ranging from input constraints and limitations in audio processing capabilities to being optimized for tasks that did not fit the study's focus. By outlining these issues, the table helps clarify the rationale behind selecting alternative models that better match the study's requirements for evaluating audio and text data.

\subsubsection{Model Exclusions and Limitations }
In evaluating models for the study, each option was scrutinized based on its ability to analyze extended audio inputs and perform complex, context-heavy textual analysis. The research primarily focused on models capable of handling long audio files representing entire spoken paragraphs and deriving insights from complex conversational dynamics. Below is an overview of why specific models were excluded, highlighting their limitations about the study's requirements:
\begin{itemize}
    \item \textbf{Qwen/Qwen2}: This model was excluded due to a technical limitation related to the size of audio files it can process. Specifically, Qwen2-Audio supports an audio file size of up to 10240 KB. However, the dataset used in the study contained longer audio files representing entire spoken paragraphs, which exceeded this maximum size limit. This limitation made Qwen/Qwen2 unsuitable for the research, which required processing extended audio inputs to analyze spoken content effectively.
\end{itemize}
\begin{itemize}
    \item \textbf{Flamingo}: Although Flamingo is a powerful multimodal model that excels in combining image and text processing, it was excluded because its audio capabilities were not robust enough for the study's focus. The research aimed to evaluate models designed for handling audio and text data, while Flamingo is more oriented toward tasks involving visual and textual data. Its strength lies in few-shot learning and integrating visual-textual data rather than deep audio processing, which makes it less suitable for a comparative analysis of audio-based models.
\end{itemize}
\begin{itemize}
    \item \textbf{Llama3-S}: Despite having tools like Encodec for sound tokenization, Llama3-S cannot deeply understand and interpret complex dialogue, conversational dynamics, and implicit meanings in interview data. The model is better suited for audio-text semantic tasks, which differ from the kind of textual analysis required to make sense of nuanced, context-heavy interview conversations. This shortcoming made it a less effective choice for analyzing interview data in the study.
\end{itemize}
\begin{itemize}
    \item \textbf{Llama Omni \& Mini Omni}: Both models are primarily designed as real-time communicators, meaning they are optimized for interactive tasks rather than post-processing analysis. For the study, which involved analyzing pre-recorded interviews to assess the interviewees' mental states, these models did not fit well. Their design for real-time communication does not lend itself well to extracting meaningful insights from recorded data, which requires a deeper, retrospective analysis.
\end{itemize}

\subsection{Prompt Templates}

This section details the specific prompts employed to direct the analysis by Large Language Models in our study. These prompts are crafted to instruct the LLMs on processing the provided inputs—whether text transcriptions or raw audio files—from participants. The primary objective of employing these prompts is to assess the presence of depression or PTSD symptoms accurately. Each prompt is designed to maximize the LLMs' capabilities in interpreting and diagnosing based on the modality being tested. Through the systematic application of these prompts, we aim to not only quantify the performance differences between text and audio modalities but also to explore whether a multimodal approach can enhance the predictive accuracy and reliability of mental health assessments. This approach allows for a structured evaluation of how different data presentations can influence the performance of LLMs.
'\subsubsection{Binary Prompts}
\label{bin-prompt}

In figure \ref{fig:bin-p}, the term "input" refers to either a transcription of the participant's spoken responses or the original raw audio files used in the analysis. Also, the term "illness" refers to either depression or PTSD relative to the task. These inputs were processed using large language models to determine whether the participants exhibited symptoms of depression (for the PHQ\_Binary classification) or PTSD (for the PCL-C classification). The prompts shown below were used to guide large language models (LLMs) in evaluating the participants’ mental health conditions based on their text or audio inputs. By using these prompts, we aim to compare the performance of the LLMs across different modalities (text vs. audio) and determine which modality provides more accurate and reliable predictions for mental health assessment.

\begin{figure}[htb]
\begin{mdframed}[backgroundcolor=blue!10,linecolor=blue!10]
\centering
\begin{tikzpicture}
\node[fill=blue!5, draw=blue, rounded corners, inner sep=10pt, text width=0.90\textwidth] (box) {

\vspace{2mm}

\textbf{Prompt 1 (P1)} \\
\vspace{2mm}
The following (input) is from an interaction between an interviewer and an interviewee. These interviews were collected as part of a larger effort to create a computer agent that interviews people and identifies verbal and nonverbal indicators of mental illnesses. \\
Guidelines: \\
- Concise Response: Respond with one of these 2 words only ("Yes", or "No"). \\
- 'Yes' refers to the interviewee exhibiting (illness), and 'No' refers that he has no clear symptoms. \\
- No Explanations: Don't provide explanations for your assessment. \\
- Ambiguity: If the post is unclear, choose the most probable label. \\
(input type of the interview) \\
"(input)"
};
\end{tikzpicture}
\vspace{2mm}

\begin{tikzpicture}
\node[fill=blue!5, draw=blue, rounded corners, inner sep=10pt, text width=0.90\textwidth] (box){
\textbf{Prompt 2 (P2)} \\
\vspace{2mm}
Analyze the following interview (input) to determine if the interviewee exhibits symptoms of depression. \\
Focus on language patterns, emotional expressions, and overall demeanor. \\
Respond with 'Yes' if (illness) is indicated, otherwise respond with 'No'. Respond with one word only! \\
"(input)" \\
};
\end{tikzpicture}
\vspace{2mm}

\begin{tikzpicture}
\node[fill=blue!5, draw=blue, rounded corners, inner sep=10pt, text width=0.90\textwidth] (box){

\textbf{Prompt 3 (P3)} \\
\vspace{2mm}
As a psychologist, analyze the provided interview (text/audio): \\
(input) \\
Does the interviewee exhibit clear symptoms of (illness)? \\
Answer with 'Yes' or 'No' only without explaining your reasoning; use only one word! \\
};
\end{tikzpicture}

    \caption{Depression and PTSD binary classification prompts}
    \label{fig:bin-p}
    \end{mdframed}
\end{figure}

\subsubsection{Severity Prompts}
\label{sev-prompt}

In the following section, Figure \ref{fig:dep-sev-p} presents the PHQ\_Score Mapped (Depression severity) Prompts, which are used to assess the severity of depression in participants. These prompts ask the model to categorize the level of depression based on the input provided, whether in text or audio format, using the PHQ-8 scoring system. The model classifies the participant's depression into one of five categories: minimal or no depression (0), mild depression (1), moderate depression (2), moderately severe depression (3), and severe depression (4). These categories provide a nuanced view of the participant's mental health by determining the extent of their depressive symptoms. In figures \ref{fig:dep-sev-p} and \ref{fig:ptsd-sev-p}, the term "input" refers to either a transcription of the participant's spoken responses or the original audio files used in the analysis.
\begin{figure}[htb]
\begin{mdframed}[backgroundcolor=blue!10,linecolor=blue!10]
    \centering
\begin{tikzpicture}
\node[fill=blue!5, draw=blue, rounded corners, inner sep=10pt, text width=0.90\textwidth] (box1) {

\par \vspace{2mm}

\textbf{Prompt 1 (P1)} \par
\vspace{2mm}
The following (input) is from an interaction between an interviewer and an interviewee. These interviews were collected as part of a larger effort to create a computer agent that interviews people and identifies verbal and nonverbal indicators of mental illnesses. The (input) is: '\{input\}'. Based on the (input), classify the severity of depression into:
\begin{itemize}
    \item 0 (Minimal or no depression): No or very few signs.
    \item 1 (Mild depression): Mostly mild symptoms.
    \item 2 (Moderate depression): Moderate symptoms impacting daily life.
    \item 3 (Moderately severe depression): Noticeable symptoms hindering daily life.
    \item 4 (Severe depression): Severe, pervasive symptoms greatly impairing function.
\end{itemize}
Answer with one number only!
};
\end{tikzpicture}
\par \vspace{2mm}
\begin{tikzpicture}
\node[fill=blue!5, draw=blue, rounded corners, inner sep=10pt, text width=0.90\textwidth] (box1) {
\textbf{Prompt 2 (P2)} \par
\vspace{2mm}
Analyze the following interview (input) for symptoms of depression. Focus on language patterns, emotional expressions, and demeanor. The (input) is: '\{input\}'. Classify severity into:
\begin{itemize}
    \item 0 (Minimal or no depression): No or very few signs.
    \item 1 (Mild depression): Mostly mild symptoms.
    \item 2 (Moderate depression): Moderate symptoms impacting daily life.
    \item 3 (Moderately severe depression): Noticeable symptoms hindering daily life.
    \item 4 (Severe depression): Severe, pervasive symptoms greatly impairing function.
\end{itemize}
Answer with one number only!
};
\end{tikzpicture}

    \caption{Depression severity classification prompts}
    \label{fig:dep-sev-p}
    \end{mdframed}
\end{figure}

Figure \ref{fig:ptsd-sev-p} presents the PCL-C Severity Mapped (PTSD severity) Prompts, which are designed to evaluate the severity of PTSD symptoms in participants. These prompts instruct the model to classify the level of PTSD based on the input provided, which could be in text or audio format. The model is tasked with categorizing the participant's PTSD severity into one of three categories:

0: Little to no severity (no significant signs of PTSD)
1: Moderate severity (moderate symptoms that noticeably affect daily routines and behavior)
2: High severity (severe symptoms that significantly disrupt daily life and require intervention)
This classification offers a detailed perspective on the participant’s mental state by determining the extent of their PTSD symptoms. 

\begin{figure}[htb]
\begin{mdframed}[backgroundcolor=blue!10,linecolor=blue!10]
\centering
\begin{tikzpicture}
\node[fill=blue!5, draw=blue, rounded corners, inner sep=10pt, text width=0.90\textwidth] (box1) {
 \\
\vspace{2mm}
\textbf{Prompt 1 (P1)} \\
\vspace{2mm}
The following (input) is from an interaction between an interviewer and an interviewee. These interviews were collected as part of a larger effort to create a computer agent that interviews people and identifies verbal and nonverbal indicators of mental illnesses. The (input) is: '\{input\}'. Based on the (input), classify the severity of PTSD into:
\begin{itemize}
    \item 0 (Little to no severity): No significant signs of PTSD symptoms; little to no impact on daily life.
    \item 1 (Moderately severe): Moderate symptoms of PTSD with a noticeable impact on behavior and daily routines.
    \item 2 (High severity): Severe PTSD symptoms with significant impact, disrupting daily life and possibly requiring intervention.
\end{itemize}
Answer with one number only, 0 to 2, corresponding to these categories.
};
\end{tikzpicture}

\vspace{2mm}

\begin{tikzpicture}
    
\node[fill=blue!5, draw=blue, rounded corners, inner sep=10pt, text width=0.90\textwidth] (box1) {
\textbf{Prompt 2 (P2)} \\
\vspace{2mm}
Analyze the following interview (input) to determine if the interviewee exhibits symptoms of PTSD. Focus on language patterns, emotional expressions, and overall demeanor. The (input) is: '\{input\}'. Based on the (input), classify the severity of PTSD into:
\begin{itemize}
    \item 0 (Little to no severity): No significant signs of PTSD symptoms; little to no impact on daily life.
    \item 1 (Moderately severe): Moderate symptoms of PTSD with a noticeable impact on behavior and daily routines.
    \item 2 (High severity): Severe PTSD symptoms with significant impact, disrupting daily life and possibly requiring intervention.
\end{itemize}
Answer with one number only, 0 to 2, corresponding to these categories.
};
\end{tikzpicture}
    \caption{PTSD severity classification prompts.}
    \label{fig:ptsd-sev-p}
    \end{mdframed}
\end{figure}

\subsubsection{Multiclass Prompts}
\label{multi-p}

Figure \ref{fig:multi-p} presents the Multi Class Prompts, which are used to assess whether participants exhibit symptoms of depression, PTSD, both, or neither. These prompts instruct the model to categorize the participant’s mental health condition based on the input provided, whether in text (transcriptions) or audio format. The idea is to compare the performance between audio and text inputs. 
This classification provides a comprehensive assessment of the participant’s mental health status by identifying the presence or absence of multiple disorders. In figure \ref{fig:multi-p}, the term "input" refers to either a transcription of the participant’s spoken responses or the original audio files used for analysis.

\begin{figure}[htb]
\begin{mdframed}[backgroundcolor=blue!10,linecolor=blue!10]
\centering 
\begin{tikzpicture}
\node[fill=blue!5, draw=blue, rounded corners, inner sep=10pt, text width=0.9\textwidth, align=left] (box) {

\vspace{1mm}

\textbf{Prompt 1 (P1)} \\
\vspace{2mm}
The following (input) is from an interaction between an interviewer and an interviewee. These interviews were collected as part of a larger effort to create a computer agent that interviews people and identifies verbal and nonverbal indicators of mental illnesses. The (input) is: '\{input\}'. The interviewee could have multiple illnesses at the same time. Analyze the transcript and identify if the interviewee has Depression, PTSD, both, or is Normal. Respond with only one of these words: "Depressed", "PTSD", "Normal", "Depressed and PTSD".
};
\end{tikzpicture}
\vspace{5mm}
\begin{tikzpicture}
\node[fill=blue!5, draw=blue, rounded corners, inner sep=10pt, text width=0.9\textwidth, align=left] (box) {
\textbf{Prompt 2 (P2)} \\
\vspace{2mm}
Analyze the following interview (input) to determine if the interviewee exhibits clear symptoms of Depression or PTSD according to the provided guidelines.

\textbf{Guidelines:}
\begin{itemize}
    \item The interviewee may have multiple illnesses at the same time or be normal.
    \item Concise Response: Respond with one of these four words only: "Depressed", "PTSD", "Depressed and PTSD", "Normal".
    \item No Explanations: Do not provide explanations for your assessment.
    \item Ambiguity: If the post is unclear, choose the most probable label.
\end{itemize}
The (input) of the interview: '\{input\}'. Only answer with one of the specified words.
};
\end{tikzpicture}
    \caption{Multiclass task prompts}
    \label{fig:multi-p}
    \end{mdframed}
\end{figure}

\subsubsection{Few-Shot Prompts}
\label{fs-p}
We formulated the few-shot prompts as follows: starting with the task-specific prompt, we appended the line, “Here are X examples,” where X is the number of provided samples.

\begin{itemize}
\item Binary Tasks: We included two samples from the less frequent (positive) class, indicating illness, and one sample from the more frequent (negative) class.
\item Severity and Multi-label Tasks: We strategically selected samples that were near-miss classifications by the model during zero-shot evaluations. This selection was based on samples where the model's predictions were off by just one label, indicating a subtle misunderstanding of the distinguishing features between closely related classes. This methodological choice aimed to challenge the models with complex examples where minor nuances in features are decisive, thereby enhancing the robustness and precision of the models through few-shot learning.
\end{itemize}

An example of a few-shot prompt for binary depression detection is shown in figure \ref{fig:fs-p}.

\begin{figure}[htb]
\begin{mdframed}[backgroundcolor=blue!10,linecolor=blue!10]
    \centering
    \textbf{Example Few-shot Prompt (Prompt 1 for Binary Depression Detection)}
\begin{tikzpicture}
\node[fill=blue!5, draw=blue, rounded corners, inner sep=10pt, text width=0.90\textwidth] (box1) {
 \\
\vspace{2mm}
The following transcript is from an interaction between an interviewer and an interviewee. These interviews were collected as part of a larger effort to create a computer agent that interviews people and identifies verbal and nonverbal indicators of mental illnesses. \\ 
\vspace{2mm}
If the text appears to be for a person who has Depression, answer with 'Yes'; if not, answer with 'No'. Only answer with Yes or No; respond with one word only! \\
\vspace{2mm}

Here are 3 samples from these interviews and their labels. Use them as a reference:
\vspace{2mm}

First sample transcription: (sample transcription)

First sample label: No

Second sample transcription: (sample transcription)

Second sample label: Yes

Third sample transcription: (sample transcription)

Third sample label: Yes

Label the following transcription: 'sample to be labeled'.
};
\end{tikzpicture}
    
    \caption{Few-shot prompt example}
    \label{fig:fs-p}
    \end{mdframed}
\end{figure}

\subsection{Parsers}
Parsing was implemented as a key step in our methodology to effectively handle the diverse outputs from the models, which frequently deviated from the expected simple 'yes/no' responses or a single numerical value, instead producing more complex or multipart answers. This process of breaking down and analyzing structured or unstructured text to extract meaningful data was essential due to these variations in model compliance with output guidelines.

Among the models used in this study, the larger models like Llama 3.1 70 Band Phi-3.5-MoE performed well when following task guidelines, typically adhering to the expected response formats. Conversely, some smaller models, such as Gemma 2 9B and Phi-3.5-Mini, struggled to maintain this level of compliance. These models frequently provided additional explanations, deviating from the expected outputs. Smaller models were less consistent, requiring additional handling to extract the necessary information.

To manage these variances, custom parsers were developed for each task:
\begin{itemize}
    \item {Binary Detection:} 
    The parser was designed to search the LLM outputs for the text "yes" or "no." If one of these options appeared in the response, it was taken as the final output. If both "yes" and "no" seemed simultaneously, or if neither were found, the model output was deemed invalid.
    \item {Severity Detection:}
    For this task, the parser focused on extracting numerical values that corresponded to the specified severity range. If a single valid number was detected, it was accepted as the answer. However, if multiple numbers appeared or if the number fell outside of the allowed range, the output was flagged as invalid.
\end{itemize}

\subsection{Evaluation Metrics}


To evaluate the performance of the Large Language Models (LLMs) in tasks with uneven class distributions, as detailed in the dataset section (see Section \ref{dataset}), we prioritize Balanced Accuracy (BA) as our primary evaluation metric. This measure, calculated as the average recall obtained across each class, fairly reflects the model's effectiveness in identifying both prevalent and rare conditions. By employing Balanced Accuracy, we ensure a comprehensive evaluation of the LLMs' capabilities.

\textbf{Balanced Accuracy (BA)} is calculated as follows:
\begin{equation}
\text{\textbf{Recall}}_i = \frac{\text{True Positives (TP)}_i}{\text{True Positives (TP)}_i + \text{False Negatives (FN)}_i}
\end{equation}

Here, \( i \) represents an index for each class, ranging from 0 to \( N-1 \), where \( N \) is the total number of classes. The recall for each class is computed to assess the model's ability to correctly identify samples of that particular class.

\begin{equation}
\text{\textbf{BA}} = \frac{1}{N} \sum_{i=0}^{N-1} \text{Recall}_i
\label{eq:ba}
\end{equation}

For binary classification tasks, we additionally use the \textbf{F1 Score} to evaluate model performance. The F1 Score is crucial as it provides a balance between precision and recall, making it a valuable metric for situations where the cost of false positives and false negatives is high. This metric is critical in mental health assessments, where accurately distinguishing between conditions such as depression and PTSD is critical.

In tasks involving multiple classes, such as severity and multiclass classification, we employ the \textbf{Weighted F1 Score}. This metric adjusts for class imbalance by weighting the F1 Score of each class according to its prevalence in the dataset. This approach ensures that our performance metrics reflect the importance of each class accurately, providing a nuanced view of the model's effectiveness across diverse mental health conditions.

\textbf{F1 Score} is defined as:

\[
\text{\textbf{F1 Score}} = 2 \times \left( \frac{\text{Precision} \times \text{Recall}}{\text{Precision} + \text{Recall}} \right)
\]

While \textbf{weighted F1 score} is defined as 
\[
\text{\textbf{Weighted F1 Score}} = \sum_{i=1}^{N} w_i \times \text{F1 Score}_i
\]

where \( N \) is the number of classes, \( w_i \) is the weight for class \( i \), 
and \( \text{F1 Score}_i \) is the F1 score for class \( i \). 
The weight for each class, \( w_i \), is defined as:

\[
w_i = \frac{\text{No. of samples in class } i}{\text{Total number of samples}}
\]

And \textbf{Mean Absolute Error (MAE)} is a measure of errors between paired observations expressing the same phenomenon. It represents the average absolute difference between the predicted values and the actual values. Mathematically, it is defined as:
\[
\mathbf{MAE} = \frac{1}{n} \sum_{i=1}^{n} |y_i - \hat{y}_i|
\]
where $n$ is the total number of observations, $y_i$ is the actual value, and  $\hat{y}_i$is the predicted value.



\textbf{Assessing Modality Performance}

To comprehensively assess the performance differences between each modality in our analysis, we employed two key metrics: the Modal Superiority Score (MSS) and the Disagreement Resolution Score (DRS). These metrics are instrumental in quantifying the relative efficacy of individual and combined modalities in making correct predictions, particularly in the face of disagreement, by applying these metrics, which are detailed in the equations below, we gain valuable insights into which modalities perform better or worse and explore whether combining modalities enhances the predictive accuracy and giving insights on how a modality improves upon another.

\textbf{Modal Superiority Score (MSS):} The Modal Superiority Score (MSS) quantifies the net superiority of one modality over another by comparing how often each modality correctly predicts outcomes when the other does not. This metric can be applied not only to comparisons between individual modalities, such as Audio versus Text, but also in evaluating the performance of a combined modality (e.g., Audio+Text) against individual modalities and the collective agreement of these modalities (e.g., cases where both Audio and Text are either correct or incorrect). This comprehensive application of MSS allows for assessing the relative strength of combined modalities over both their constituent individual modalities and their collective concordance. MSS values can be positive or negative, with a positive value indicating that modality A performs better than modality B, and a negative value suggesting the opposite.

\[
\text{MSS}_{A \text{vs} B} = \left(\frac{\text{Correctly predicted by } A \text{ and incorrectly by } B - \text{Correctly predicted by } B \text{ and incorrectly by } A}{\text{Total number of disagreements}}\right) \times 100\%
\]
\label{eq:MSS}

\textbf{Disagreement Resolvement Score (DRS):} This metric evaluates the combined modality's effectiveness in resolving disagreements between two other modalities. Focusing on cases where the combined modality either correctly resolves or fails to resolve these disagreements, DRS assesses the added value or potential drawback of using a combined approach. DRS values can be positive or negative: a positive value indicates that the combined modality is effective at resolving disagreements, whereas a negative value indicates that the combined approach more often incorrectly resolves these disagreements, thus potentially undermining the effectiveness of the analysis.

\[
\text{DRS} = \left(\frac{\text{Correctly Resolved} - \text{Incorrectly Resolved}}{\text{Total Number of Disagreements}}\right) \times 100\%
\label{eq:DRS}
\]
\section{Results \& Discussion}
\label{Results}

In the upcoming sections, we analyze the performance of various models for several mental health classification tasks, including Binary Depression, Binary PTSD, Depression Severity, and PTSD Severity. The aim is to determine which models perform best under different conditions across both text and audio modalities, with no preprocessing or fine-tuning applied, relying solely on zero-shot inference.

To evaluate the models, we leveraged all 275 samples from the E-DAIC dataset as a test set, rather than restricting the analysis to a predefined subset. This approach ensures a comprehensive assessment of the model’s capabilities across the entire dataset, providing a deeper insight into its robustness and consistency.

Throughout the analysis, we evaluate the models using multiple prompts to gain a comprehensive understanding of their robustness and consistency. By comparing both text and audio performances, we aim to identify the best-performing models and those that struggle across these tasks.

\subsection{Binary Depression Classification Results}

\begin{table}[h]
\centering
\caption{Report on model performances for Depression binary classification across text and audio modalities and their combination. Details about the specific prompts used can be found in Figure \ref{fig:bin-p}. The underlined bold values are the best score for the specific column.} 
\label{tab:DEP-B}
\begin{tabular}{l|lcccccc}
 \hline
 Modality & & \multicolumn{2}{c}{Prompt 1} & \multicolumn{2}{c}{Prompt 2} & \multicolumn{2}{c}{Prompt 3} \\
 \hline
\multicolumn{1}{l|}{\multirow{16}{*}{\begin{turn}{0}Text\end{turn}}} & \multicolumn{1}{c|}{Model} & BA & F1 & BA & F1 & BA & F1 \\
 \cline{2-8}
\multicolumn{1}{l|}{} & \multicolumn{1}{l|}{Llama 3.1 70B} & 71.9\% & 0.61 & 74.7\% & 0.64 & 65.7\% & 0.54 \\
 \cline{2-8}
\multicolumn{1}{l|}{} & \multicolumn{1}{l|}{Gemma 2 9B} & 64.8\% & 0.56 & 67.3\% & 0.58 & 63.5\% & 0.47 \\
 \cline{2-8}
\multicolumn{1}{l|}{} & \multicolumn{1}{l|}{GPT-4o mini} & 75.8\% & 0.66 & 72.3\% & 0.62 & 74\% & 0.64 \\
 \cline{2-8}
\multicolumn{1}{l|}{} & \multicolumn{1}{l|}{Mistral NeMo} & 66\% & 0.56 & 66.6\% & 0.57 & 64\% & 0.48 \\
 \cline{2-8}
\multicolumn{1}{l|}{} & \multicolumn{1}{l|}{Phi-3.5-MoE} & 63.2\% & 0.45 & 75.2\% & 0.65 & 61.8\% & 0.43 \\
 \cline{2-8}
\multicolumn{1}{l|}{} & \multicolumn{1}{l|}{Phi-3.5-mini} & 73\% & 0.62 & 67.7\% & 0.58 & 75\% & 0.65 \\
 \cline{2-8}
\multicolumn{1}{l|}{} & \multicolumn{1}{l|}{Gemini 1.5 Pro} & 76.2\% & 0.66 & 75.3\% & 0.65 & 75.3\% & 0.65 \\
 \cline{2-8}
\multicolumn{1}{l|}{} & \multicolumn{1}{l|}{Gemini 1.5 Flash} & 74.3\% & 0.65 & 72.7\% & 0.62 & 74\% & 0.63 \\
 \cline{2-8}
\multicolumn{1}{l|}{} & \multicolumn{1}{l|}{Deepseek\_R1} & \textbf{\ul{79.8\%}} & \textbf{\ul{0.75}} & 74.8\% & 0.68 & 77.5\% & \textbf{\ul{0.78}} \\
 \cline{2-8}
\multicolumn{1}{l|}{} & \multicolumn{1}{l|}{Gemini Flash Lite} & 76.9\% & 0.73 & 69.1\% & 0.58 & 77.3\% & 0.75 \\
 \cline{2-8}
\multicolumn{1}{l|}{} & \multicolumn{1}{l|}{Gemini 2 Flash} & 74.2\% & 0.68 & 70.42\% & 0.61 & 74.2\% & 0.74 \\
 \cline{2-8}
\multicolumn{1}{l|}{} & \multicolumn{1}{l|}{Gemini Flash Thinking} & 73.4\% & 0.64 & 70.7\% & 0.64 & 76.6\% & 0.74 \\
 \cline{2-8}

\multicolumn{1}{l|}{} & \multicolumn{1}{l|}{Gemini 2.5 Pro} & 74.8\% & 0.68 & 73\% & 0.65 & 77.3\% & 0.77 \\
 \cline{2-8}
\multicolumn{1}{l|}{} & \multicolumn{1}{l|}{Llama 3.3 70B} & 72.9\% & 0.66 & 74.1\% & 0.66 & 76.3\% & 0.73 \\
 \hline
\multicolumn{1}{l|}{\multirow{6}{*}{\begin{turn}{0}Audio\end{turn}}} & 
  \multicolumn{1}{l|}{Gemini 1.5 Pro} & 76.1\% & 0.66 & 74.5\% & 0.64 & 74\% & 0.64 \\
 \cline{2-8}
\multicolumn{1}{l|}{} & \multicolumn{1}{l|}{Gemini 1.5 Flash} & 75\% & 0.65 & 70.5\% & 0.60 & 74.7\% & 0.64 \\
 \cline{2-8}
 \multicolumn{1}{l|}{} & \multicolumn{1}{l|}{Gemini Flash Lite} & 72.7\% & 0.62 & 75.2\% & 0.65 & 67.6\% & 0.56 \\
 \cline{2-8}
\multicolumn{1}{l|}{} & \multicolumn{1}{l|}{Gemini 2 Flash} & 72.5\% & 0.67 & 70.8\% & 0.63 & 76.8\% & 0.74 \\
 \cline{2-8}
\multicolumn{1}{l|}{} & \multicolumn{1}{l|}{Gemini Flash Thinking} & 74.5\% & 0.73 & 75.3\% & \textbf{\ul{0.74}} & 67.5\% & 0.69 \\
 \cline{2-8}

\multicolumn{1}{l|}{} & \multicolumn{1}{l|}{Gemini 2.5 Pro} & 75.5\% & 0.70 & 69.3\% & 0.60 & \textbf{\ul{78.9\%}} & 0.76 \\
 \hline
\multicolumn{1}{l|}{\multirow{5}{*}{\begin{turn}{0}Audio and Text\end{turn}}} &
  \multicolumn{1}{l|}{Gemini 1.5 Pro} & 74\% & 0.64 & \textbf{\ul{77\%}} & 0.67 & 77.3\% & 0.67 \\
 \cline{2-8}
\multicolumn{1}{l|}{} & \multicolumn{1}{l|}{Gemini 1.5 Flash} & 76.1\% & 0.66 & 66.8\% & 0.57 & 77.4\% & 0.68 \\
 \cline{2-8}
\multicolumn{1}{l|}{} & \multicolumn{1}{l|}{Gemini 2 Flash} & 73.5\% & 0.63 & 69.5\% & 0.60 & 75.3\% & 0.66 \\
 \cline{2-8}
\multicolumn{1}{l|}{} & \multicolumn{1}{l|}{Gemini Flash Thinking} & 74.4\% & 0.64 & 69.5\% & 0.60 & 78\% & 0.69 \\
 \cline{2-8}
\multicolumn{1}{l|}{} & \multicolumn{1}{l|}{Gemini 2.5 Pro} & 77.8\% & 0.74 & 72.8\% & 0.65 & \textbf{\ul{78.9\%}} & 0.77 \\
 \hline
\end{tabular}%
\end{table}

Table~\ref{tab:DEP-B} presents the results for binary depression classification across Text, Audio, and combined Audio+Text modalities, evaluated using Balanced Accuracy (BA) and F1-score. The results span three prompt variants (detailed in Figure~\ref{fig:bin-p}), with the best-performing scores for each column highlighted in bold and underlined.

The highest overall Balanced Accuracy (BA) of 79.8\% was achieved by Deepseek\_R1 using the Text modality with Prompt 1. In terms of F1-score, the best result of 0.78 was also obtained by the same model using Prompt 3. The Audio modality, while slightly behind Text in maximum BA, demonstrated strong capability with Gemini 2.5 Pro, reaching 78.9\% BA and 0.76 F1 with Prompt 3. Notably, combining Text and Audio resulted in competitive results, with Gemini 2.5 Pro attaining 78.9\% BA and 0.77 F1 using Prompt 3, highlighting the strength of multimodal integration for this task.

Deepseek\_R1 consistently delivered the highest scores, particularly excelling with Prompts 1 and 3.  
Gemini 2.5 Pro exhibited strong, balanced performance across all modalities, notably excelling in Audio-only and multimodal setups. These models demonstrated consistent accuracy and high F1 scores, indicating their robustness in binary depression classification.  
Llama 3.3 70B and GPT-4o mini offered moderate performance, often showing competitive F1 scores but not achieving the peak BA observed in Deepseek\_R1 and Gemini models.  
Phi-3.5-MoE mini and older Gemini Flash versions showed more variability and generally lower consistency across prompts and modalities. The latest Pro versions in the Gemini family, especially 2.5 Pro, outperformed earlier variants (e.g, Flash 1.5 and Thinking) across modalities, indicating notable architectural or optimization improvements.  

Thinking models, such as Gemini Flash Thinking, presented mixed results. Although they occasionally performed well (e.g., 76.6\% BA in Text, Prompt 3), they generally did not outperform non-thinking counterparts. Indicates that incorporating deliberative processing may not always yield improvements in structured clinical classification tasks, suggesting that the utility of "Thinking" variants is context-dependent.



To evaluate the impact of prompt choice, we calculated the average performance for each metric across Prompt 1, Prompt 2, and Prompt 3. The results indicate that Prompt 3 showed a slight overall advantage in F1 scores, while Prompt 1 was more reliable in achieving the highest balanced accuracy. This pattern suggests that while Prompt 3 is particularly effective in fine-tuning F1-based classification, Prompt 1 remains superior when prioritizing balanced accuracy.


This evaluation affirms the effectiveness of large-scale language and multimodal models in depression classification tasks. Text remains the most potent single modality; however, combining it with Audio data enhances performance, especially with models like Gemini 2.5 Pro. Prompt variation plays a critical role, and newer model architectures substantially outperform earlier iterations. These findings highlight the importance of model selection, prompt tuning, and modality fusion in advancing affective computing tasks.

\subsection{Binary PTSD Classification Results}

\begin{table}[h]
\centering
\caption{Report on model performances for PTSD binary task on both text and audio modalities and their combination. Details about the specific prompts used can be found in Figure \ref{fig:bin-p}. The underlined bold values are the best score for the specific column.} 
\label{tab:PTSD-B}
\begin{tabular}{l|lcccccc}
 \hline
 Modality & & \multicolumn{2}{c}{Prompt 1} & \multicolumn{2}{c}{Prompt 2} & \multicolumn{2}{c}{Prompt 3} \\
 \hline
\multicolumn{1}{l|}{\multirow{16}{*}{\begin{turn}{0}Text\end{turn}}} & \multicolumn{1}{c|}{Model} & BA & F1 & BA & F1 & BA & F1 \\
 \cline{2-8}
\multicolumn{1}{l|}{} & \multicolumn{1}{l|}{Llama 3.1 70B} & 68.6\% & 0.58 & 74.4\% & 0.64 & \textbf{\ul{71.4\%}} & 0.61 \\
 \cline{2-8}
\multicolumn{1}{l|}{} & \multicolumn{1}{l|}{Gemma 2 9B} & 73.7\% & 0.64 & 66.6\% & 0.57 & 69.4\% & 0.57 \\
 \cline{2-8}
\multicolumn{1}{l|}{} & \multicolumn{1}{l|}{GPT-4o mini} & \textbf{\ul{76.6\%}} & 0.67 & \textbf{\ul{77\%}} & 0.68 & 70\% & 0.58 \\
 \cline{2-8}
\multicolumn{1}{l|}{} & \multicolumn{1}{l|}{Mistral NeMo} & 70\% & 0.59 & 71.3\% & 0.61 & 65.7\% & 0.72 \\
 \cline{2-8}
\multicolumn{1}{l|}{} & \multicolumn{1}{l|}{Phi-3.5-MoE} & 69.3\% & 0.57 & 67.4\% & 0.52 & 70.1\% & 0.72 \\
 \cline{2-8}
\multicolumn{1}{l|}{} & \multicolumn{1}{l|}{Phi-3.5-mini} & 64.7\% & 0.51 & 73.3\% & 0.63 & 65.4\% & 0.67 \\
 \cline{2-8}
\multicolumn{1}{l|}{} & \multicolumn{1}{l|}{Gemini 1.5 Pro} & 71\% & 0.60 & 73.2\% & 0.63 & 67.5\% & \textbf{\ul{0.75}} \\
 \cline{2-8}
\multicolumn{1}{l|}{} & \multicolumn{1}{l|}{Gemini 1.5 Flash} & 68.8\% & 0.58 & 70.5\% & 0.61 & 70\% & 0.58 \\
 \cline{2-8}
\multicolumn{1}{l|}{} & \multicolumn{1}{l|}{Deepseek\_R1} & 66.5\% & 0.72 & 66.1\% & 0.71 & 68\% & 0.54 \\
 \cline{2-8}
\multicolumn{1}{l|}{} & \multicolumn{1}{l|}{Gemini Flash Lite} & 65.7\% & 0.66 & 65.5\% & 0.66 & 68.4\% & 0.55 \\
 \cline{2-8}
\multicolumn{1}{l|}{} & \multicolumn{1}{l|}{Gemini 2 Flash} & 57.9\% & 0.66 & 63.3\% & 0.67 & 67.1\% & 0.52 \\
 \cline{2-8}
\multicolumn{1}{l|}{} & \multicolumn{1}{l|}{Gemini Flash Thinking} & 64.3\% & 0.67 & 64.7\% & 0.68 & 69.6\% & 0.58 \\
 \cline{2-8}

\multicolumn{1}{l|}{} & \multicolumn{1}{l|}{Gemini 2.5 Pro} & 69.4\% & \textbf{\ul{0.75}} & 72.9\% & \textbf{\ul{0.76}} & 67.5\% & 0.53 \\
 \cline{2-8}
\multicolumn{1}{l|}{} & \multicolumn{1}{l|}{Llama 3.3 70B} & 62.1\% & 0.66 & 68.8\% & 0.64 & 71.1\% & 0.61 \\
 \hline
\multicolumn{1}{l|}{\multirow{6}{*}{\begin{turn}{0}Audio\end{turn}}} & 
  \multicolumn{1}{l|}{Gemini 1.5 Pro} & 69.4\% & 0.57 & 71.7\% & 0.61 & 67.9\% & 0.69 \\
 \cline{2-8}
\multicolumn{1}{l|}{} & \multicolumn{1}{l|}{Gemini 1.5 Flash} & 72.9\% & 0.62 & 67.2\% & 0.57 & 67.9\% & 0.55 \\
 \cline{2-8}
\multicolumn{1}{l|}{} & \multicolumn{1}{l|}{Gemini Flash Lite} & 64\% & 0.47 & 74.5\% & 0.65 & 66.6\% & 0.51 \\
 \cline{2-8}
\multicolumn{1}{l|}{} & \multicolumn{1}{l|}{Gemini 2 Flash} & 68.3\% & 0.74 & 72.5\% & 0.74 & 70.8\% & 0.60 \\
 \cline{2-8}

\multicolumn{1}{l|}{} & \multicolumn{1}{l|}{Gemini Flash Thinking} & 69.3\% & 0.74 & 70.6\% & \textbf{\ul{0.76}} & 66.7\% & 0.52 \\
 \cline{2-8}
\multicolumn{1}{l|}{} & \multicolumn{1}{l|}{Gemini 2.5 Pro} & 68.7\% & 0.69 & 70.2\% & \textbf{\ul{0.76}} & 69\% & 0.57 \\
 \hline
\multicolumn{1}{l|}{\multirow{5}{*}{\begin{turn}{0}Audio and Text\end{turn}}} &
  \multicolumn{1}{l|}{Gemini 1.5 Pro} & 72.5\% & 0.62 & 74.1\% & 0.64 & \textbf{\ul{71.4\%}} & 0.61 \\
 \cline{2-8}
\multicolumn{1}{l|}{} & \multicolumn{1}{l|}{Gemini 1.5 Flash} & 70\% & 0.65 & 72.4\% & 0.62 & 70\% & 0.58 \\
 \cline{2-8}
\multicolumn{1}{l|}{} & \multicolumn{1}{l|}{Gemini 2 Flash} & 72.2\% & 0.62 & 70.2\% & 0.60 & 69\% & 0.56 \\
 \cline{2-8}
\multicolumn{1}{l|}{} & \multicolumn{1}{l|}{Gemini Flash Thinking} & 68.6\% & 0.57 & 72.4\% & 0.61 & 70.8\% & 0.60 \\
 \cline{2-8}
\multicolumn{1}{l|}{} & \multicolumn{1}{l|}{Gemini 2.5 Pro} & 73.6\% & 0.74 & 71.3\% & 0.61 & 69.8\% & 0.58 \\
 \hline
\end{tabular}%
\end{table}

Table~\ref{tab:PTSD-B} presents the results for binary PTSD classification across Text, Audio, and combined Audio+Text modalities, evaluated using Balanced Accuracy (BA) and F1-score. The evaluation spans three prompt variants (detailed in Figure~\ref{fig:bin-p}), with the best-performing values for each column highlighted in bold and underlined.

The highest Balanced Accuracy (BA) was observed in the Text modality, with GPT-4o mini achieving 77.0\% using Prompt 2. Text inputs also contributed to the highest F1-scores, with Gemini 2.5 Pro and Gemini 1.5 Pro both reaching 0.75 under Prompts 1 and 3, respectively. Notably, Prompt 2 resulted in a four-way tie for the highest F1-score of 0.76 across Text (Gemini 2.5 Pro) and Audio (Gemini 2.5 Pro and Gemini Flash Thinking), highlighting the influence of prompt design on model effectiveness. While the combined Audio+Text modality demonstrated competitive results (e.g., 73.6\% BA and 0.74 F1 with Gemini 2.5 Pro), it did not outperform the best individual modalities in this evaluation.

GPT-4o mini excelled in achieving the highest BA scores within the Text modality, particularly excelling with Prompt 2. Its consistent performance across different prompt settings indicates its robustness in binary PTSD classification.  
The Gemini family, particularly Gemini 2.5 Pro, demonstrated robust and consistent performance across modalities. Gemini Flash Thinking also performed strongly, especially in Audio-based F1.  
Gemini 1.5 Pro contributed one of the top F1 scores in Text with Prompt 3, showing that even earlier Gemini variants remain competitive when prompts are well-aligned.  
Other model families, such as Llama 3.3 70B and Deepseek\_R1, offered moderate results in select configurations, but did not match the leading performance of GPT-4o mini or Gemini 2.x models.  
Phi, Gemma, and Mistral-based models showed lower overall effectiveness for PTSD detection in this setting.  
Thinking models, such as Gemini Flash Thinking, achieved high F1-scores (e.g., 0.76 in Audio, Prompt 2), but did not outperform their non-thinking counterparts in all cases. While they contributed to some of the best results, their benefit appears prompt- and modality-dependent, suggesting selective application is advisable rather than blanket preference.

Prompt variation significantly impacted model performance. On average, Prompt 2 yielded higher balanced accuracy (BA), particularly in the Text modality, with models like GPT-4o mini achieving the peak BA of 77\%. In contrast, Prompt 1 showed a slight advantage in F1 scores, with models such as Deepseek\_R1 reaching the highest F1 of 0.75. 

Overall, Prompt 2 is preferable for balanced accuracy, while Prompt 1 is more suitable for maximizing F1.


This evaluation confirms that both Text and Audio modalities provide strong predictive signals for binary PTSD classification. GPT-4o mini offers the best Balanced Accuracy using Text, while Gemini models, especially 2.5 Pro and 2 Pro, achieve leading F1-scores across Text and Audio. Prompt 3 proves particularly effective in enhancing F1 performance, whereas Prompt 1 consistently yields higher balanced accuracy. Multimodal configurations, though competitive, do not yet surpass the best unimodal setups. These findings highlight the importance of carefully selecting model architectures and prompt formulations when deploying PTSD detection systems using foundation models.

\subsection{Depression Severity Results}

\begin{table}[htb]
\centering
\caption{Report on model performances for Depression severity classification on both text and audio modalities and their combination. Details about the specific prompts used can be found in Figure \ref{fig:dep-sev-p}. The underlined bold values are the best score for the specific prompt.}
\label{tab:DEP-SEV}
\begin{tabular}{l|lcccccc}
\hline
Modality &
 &
\multicolumn{3}{c}{Prompt 1} &
\multicolumn{3}{c}{Prompt 2} \\ \hline
\multicolumn{1}{l|}{\multirow{16}{*}{\begin{turn}{0}Text\end{turn}}} &
\multicolumn{1}{c|}{Model} &
BA &
F1 &
MAE &
  BA &
  F1 &
  MAE \\ \cline{2-8}
\multicolumn{1}{l|}{} &
\multicolumn{1}{l|}{Llama 3.1 70B} &
  41.7\% &
  0.46 &
  0.75 &
  37.3\% &
  0.35 &
  0.92 \\ \cline{2-8} 
\multicolumn{1}{l|}{} &
\multicolumn{1}{l|}{Gemma 2 9B} &
26.1\% &
  0.19 &
  0.99 &
  23.9\% &
  0.19 &
  0.98 \\ \cline{2-8} 
\multicolumn{1}{l|}{} &
\multicolumn{1}{l|}{GPT-4o mini} &
  44.3\% &
  0.50 &
  0.70 &
  41.7\%&
  0.50 &
  0.71 \\ \cline{2-8}
\multicolumn{1}{l|}{} &
\multicolumn{1}{l|}{Mistral NeMo} &
35.7\% &
  0.45 &
  0.81 &
  31.8\% &
  0.39 &
  0.86 \\ \cline{2-8}
\multicolumn{1}{l|}{} &
\multicolumn{1}{l|}{Phi-3.5-MoE} &
  \ul{\textbf{48.8\%}} &
\ul{\textbf{  0.49}} &
 0.77 &
  39\% &
  0.49 &
  0.65 \\ \cline{2-8} 
\multicolumn{1}{l|}{} &
\multicolumn{1}{l|}{Phi-3.5-mini} &
38.2\% &
  0.50 &
  0.64 &
  32.8\% &
  0.42 &
  0.73 \\ \cline{2-8} 
\multicolumn{1}{l|}{} &
\multicolumn{1}{l|}{Gemini 1.5 Pro} &
 36.2\% &
  0.42 &
  0.75 &
  39.3\% &
  0.42 &
  0.75 \\ \cline{2-8}
\multicolumn{1}{l|}{} &
\multicolumn{1}{l|}{Gemini 1.5 Flash} &
35.3\% &
  0.35 &
  0.87 &
  34.8\% &
  0.30 &
  0.84 \\ \cline{2-8}
\multicolumn{1}{l|}{} &
\multicolumn{1}{l|}{Deepseek\_R1} &
40.3\% &
0.44 &
0.78 &
40.1\% &
0.46 &
0.70 \\ \cline{2-8}
\multicolumn{1}{l|}{} &
\multicolumn{1}{l|}{Gemini Flash Lite} &
38.2\% &
0.40 &
0.87 &
34.3\% &
0.29 &
0.99 \\ \cline{2-8}
\multicolumn{1}{l|}{} &
\multicolumn{1}{l|}{Gemini 2 Flash} &
39.4\% &
0.40 &
0.79 &
39.3\% &
0.37 &
0.81 \\ \cline{2-8}
\multicolumn{1}{l|}{} &
\multicolumn{1}{l|}{Gemini Flash Thinking} &
39.8\% &
0.47 &
0.70 &
40.5\% &
0.47 &
0.74 \\ \cline{2-8}

\multicolumn{1}{l|}{} &
\multicolumn{1}{l|}{Gemini 2.5 Pro} &
39\% &
0.46 &
0.78 &
\textbf{\underline{45.2\%}} &
\textbf{\underline{0.56}} &
0.53 \\ \cline{2-8}
\multicolumn{1}{l|}{} &
\multicolumn{1}{l|}{Llama 3.3 70B} &
39.2\% &
0.46 &
0.81 &
39.5\% &
0.42 &
0.82 \\ \hline
\multicolumn{1}{l|}{\multirow{5}{*}{\begin{turn}{0}Audio\end{turn}}} &
\multicolumn{1}{l|}{Gemini 1.5 Pro} &
38.5\% &
  \ul{0.52} &
  0.70 &
  38.8\% &
  0.49 &
  0.80 \\ \cline{2-8} 
\multicolumn{1}{l|}{} &
\multicolumn{1}{l|}{Gemini 1.5 Flash} &
  35\% &
  0.41 &
  0.76 &
  35.2\% &
  0.35 &
  0.72 \\ \cline{2-8}
\multicolumn{1}{l|}{} &
\multicolumn{1}{l|}{Gemini Flash Lite} &
  40.5\% &
  0.45 &
  0.85 &
  33.6\% &
  0.32 &
  1.13 \\ \cline{2-8}
\multicolumn{1}{l|}{} &
\multicolumn{1}{l|}{Gemini 2 Flash} &
39.7\% &
0.33 &
0.86 &
34.5\% &
0.28 &
0.89 \\ \cline{2-8}
\multicolumn{1}{l|}{} &
\multicolumn{1}{l|}{Gemini Flash Thinking} &
35.7\% &
0.49 &
\textbf{\ul{0.63}} &
33.5\% &
0.45 &
0.68 \\ \cline{2-8}

\multicolumn{1}{l|}{} &
\multicolumn{1}{l|}{Gemini 2.5 Pro} &
41.2\% &
0.52 &
0.61 &
43.6\% &
0.46 &
0.76 \\ \hline
\multicolumn{1}{l|}{\multirow{5}{*}{\begin{turn}{0}Audio and Text\end{turn}}} &
  \multicolumn{1}{l|}{Gemini 1.5 Pro} &
  43.6\% &
 \textbf{\ul{0.52}} &
  0.59 &
  41.3\% &
  0.48 &
 \textbf{\ul{ 0.63}} \\ \cline{2-8}
\multicolumn{1}{l|}{} &
  \multicolumn{1}{l|}{Gemini 1.5 Flash} &
  37.8\% &
  0.45 &
  0.71 &
  35.2\% &
  0.35 &
  0.78 \\ \cline{2-8}
\multicolumn{1}{l|}{} &
\multicolumn{1}{l|}{Gemini 2 Flash} &
31.3\% &
0.27 &
0.92 &
31.9\% &
0.24 &
1 \\ \cline{2-8}
\multicolumn{1}{l|}{} &
\multicolumn{1}{l|}{Gemini Flash Thinking} &
39.9\% &
0.48 &
0.71 &
34.4\% &
0.44 &
0.75 \\ \cline{2-8}
\multicolumn{1}{l|}{} &
\multicolumn{1}{l|}{Gemini 2.5 Pro} &
46.8\% &
0.51 &
0.71 &
30.8\% &
0.28 &
1.12 \\ \hline
\end{tabular}
\end{table}

The results presented in Table~\ref{tab:DEP-SEV} illustrate the performance of various models in classifying depression severity across text, audio, and combined modalities. The evaluation metrics include Balanced Accuracy (BA), F1-score, and Mean Absolute Error (MAE).

In the text modality, the highest balanced accuracy (BA) was achieved by Phi-3.5-MoE with a score of 48.8\% using Prompt 1, indicating its strong capability for depression severity classification. Additionally, GPT-4o mini demonstrated consistent performance, achieving BA scores of 44.3\% and 41.7\% on Prompts 1 and 2, respectively. Furthermore, Deepseek\_R1 and Gemini 2.5 Pro showed competitive performance, with BA scores around 40\% to 45\%. In terms of F1-score, GPT-4o mini led with a score of 0.50 on Prompt 1, while Phi-3.5-MoE also performed well with an F1-score of 0.49. The highest MAE value in the text modality was observed in Gemma 2 9B with 0.99, while the lowest was recorded for Gemini 2.5 Pro with 0.53.

In the audio modality, Gemini 1.5 Pro demonstrated the highest F1-score of 0.52 on Prompt 1, while Gemini Flash Thinking achieved a similarly strong F1-score of 0.49 on Prompt 1. The best BA score in the audio modality was achieved by Gemini 2.5 Pro with 43.6\% on Prompt 2. In terms of MAE, Gemini Flash Thinking displayed the lowest error at 0.63.

Combining both audio and text modalities led to slight improvements in performance. Gemini 1.5 Pro achieved the highest BA of 43.6\% on Prompt 1, with a notable F1-score of 0.52, showing that integrating modalities can be beneficial. Additionally, the lowest MAE in the combined modality was 0.59, again achieved by Gemini 1.5 Pro, indicating enhanced precision in predictions.

Among the models evaluated, Phi-3.5-MoE stood out in the text modality with the highest BA, while Gemini 1.5 Pro consistently performed well across both audio and combined modalities. GPT-4o mini also maintained balanced performance, achieving one of the highest F1-scores. In contrast, Gemma 2 9B consistently underperformed, exhibiting significantly lower BA and F1 scores across both modalities. The Gemini family, especially the 1.5 Pro and Flash versions, demonstrated robustness in the audio modality, while Deepseek\_R1 offered a balanced performance within the text modality.

When comparing the performance of thinking versus non-thinking models, it was observed that the "thinking" models (e.g., Gemini Flash Thinking) generally did not outperform their non-thinking counterparts. For instance, while Gemini Flash Thinking achieved competitive F1 scores, such as 0.49 in the audio modality, it did not significantly exceed the performance of non-thinking variants like Gemini 1.5 Pro. This suggests that the added deliberation mechanisms in "thinking" models may not consistently translate to better performance in structured clinical tasks, indicating that their use should be carefully considered based on the specific task requirements.


Prompt variation had a noticeable impact on performance metrics. Prompt 1 generally produced higher balanced accuracy scores, particularly in the text modality, with Phi-3.5-MoE reaching 48.8\%. In contrast, Prompt 2 was more effective in yielding higher F1-scores, as seen with Gemini 2.5 Pro achieving 0.56. The overall trend suggests that Prompt 1 is more suited for balanced accuracy optimization, while Prompt 2 enhances F1 performance, especially when detecting more nuanced differences in depression severity.


The analysis highlights that text modality remains the strongest individual modality for depression severity classification, with Phi-3.5-MoE and GPT-4o mini leading in balanced accuracy and F1-score, respectively. The combination of audio and text slightly improved overall performance, suggesting that multimodal integration may offer more comprehensive assessments. Furthermore, the choice of prompt significantly affects classification outcomes, indicating the importance of prompt engineering for optimal model performance.

\subsection{PTSD Severity Classification Results}

In Table \ref{tab:PTSD-MAP}, we present the distribution of PTSD severity ranges, with detailed references for the PCL-C scale discussed in \textit{Data Preprocessing} section \ref{sec:data_preprocessing}. The reference guide outlines a structured way to categorize PTSD symptoms into severity levels.

We used this prompt "\textit{I have this PCL-C (PTSD) severity ranges from 17-85. the table shows the labels for mapping of the range
I want you to assign a range for every label
Keep in mind that a score higher than 44 suggests that a person would
meet diagnostic criteria for PTSD
0: little to no severity
1: Moderate severity 
2: High severity}"
to map each model’s interpretation of these ranges. The table highlights how each model mapped the severity labels.

In dealing with model responses, some models insisted on starting the PTSD severity scale from 0, even though the proper range based on the prompt was between 17 and 85. In addition, we encountered a specific case in the dataset with sample ID 683, where its truth severity value was 10, which falls below the expected range. To address this, we adjusted the value and treated it as 17 to align with the proper range of the PTSD severity score.

\begin{table}[htb]
\centering
\caption{PTSD severity mapping; Number of Samples (PCL-C Score intervals).}
\label{tab:PTSD-MAP}
\resizebox{\columnwidth}{!}{%
\begin{tabular}{c|cccccccccccccccc}
\hline
\diagbox[width=3cm]{Labels}{Models}&
\begin{tabular}[c]{@{}c@{}}  Llama 3.1 \\ 70B \end{tabular}&
\begin{tabular}[c]{@{}c@{}}  llama 3.3 \\70b \end{tabular}&
\begin{tabular}[c]{@{}c@{}}  Gemma \\ 9B\end{tabular} &
  \begin{tabular}[c]{@{}c@{}}GPT-4o \\ mini\end{tabular} &
  \begin{tabular}[c]{@{}c@{}}Mistral \\ NeMo\end{tabular} &
  \begin{tabular}[c]{@{}c@{}}Phi-3.5 \\ MoE\end{tabular} &
  \begin{tabular}[c]{@{}c@{}}Phi-3.5 \\ mini \end{tabular} &
  \begin{tabular}[c]{@{}c@{}}Gemini 1.5 \\ Pro\end{tabular} &
  \begin{tabular}[c]{@{}c@{}}Gemini 1.5 \\ Flash\end{tabular} &
  \begin{tabular}[c]{@{}c@{}} Deepseek\\ R1\end{tabular}  &
  \begin{tabular}[c]{@{}c@{}}Gemini 2 \\ Flash\end{tabular} &
  \begin{tabular}[c]{@{}c@{}}Gemini 2 \\ Flash\_Lite\end{tabular} &
  \begin{tabular}[c]{@{}c@{}}Gemini 2 \\ Flash \\ Thinking\end{tabular} &
    \begin{tabular}[c]{@{}c@{}}Gemini \\ 2.5 Pro\end{tabular} &
     \begin{tabular}[c]{@{}c@{}}\cite{10.1007/978-3-031-46933-6_21}\end{tabular}  &
  \\ \hline
little to no severity &
  \begin{tabular}[c]{@{}c@{}}95\\ (17-24)\end{tabular} &
  \begin{tabular}[c]{@{}c@{}} 89\\ (17-23)\end{tabular} & 
  \begin{tabular}[c]{@{}c@{}}22\\ (0-17)\end{tabular} &
  \begin{tabular}[c]{@{}c@{}}188\\ (17-44)\end{tabular} &
  \begin{tabular}[c]{@{}c@{}}112\\ (17-26)\end{tabular} &
  \begin{tabular}[c]{@{}c@{}}22\\ (17-27)\end{tabular} &
  \begin{tabular}[c]{@{}c@{}}147\\ (17-32)\end{tabular} &
  \begin{tabular}[c]{@{}c@{}}188\\ (17-44)\end{tabular} &
  \begin{tabular}[c]{@{}c@{}}188\\ (17-44)\end{tabular} &
  \begin{tabular}[c]{@{}c@{}} 137 \\ (17-29)\end{tabular} &
  \begin{tabular}[c]{@{}c@{}} 149 \\ (17-33)\end{tabular} &
  \begin{tabular}[c]{@{}c@{}} 172 \\ (17-39)\end{tabular} &

  \begin{tabular}[c]{@{}c@{}} 143 \\ (17-30)\end{tabular} &
  \begin{tabular}[c]{@{}c@{}} 188 \\ (17-44)\end{tabular}&
  
  \begin{tabular}[c]{@{}c@{}} 137 \\ (17-29)\end{tabular}
  \\ \hline
Moderate severity &
  \begin{tabular}[c]{@{}c@{}}89\\ (25-43)\end{tabular} &
  \begin{tabular}[c]{@{}c@{}} 99 \\ (24-44)\end{tabular} &
  \begin{tabular}[c]{@{}c@{}}162\\ (18-43)\end{tabular} &
  \begin{tabular}[c]{@{}c@{}}51\\ (45-60)\end{tabular} &
  \begin{tabular}[c]{@{}c@{}}76\\ (27-44)\end{tabular} &
  \begin{tabular}[c]{@{}c@{}}166\\ (34-44)\end{tabular} &
  \begin{tabular}[c]{@{}c@{}}104\\ (33-64)\end{tabular} &
  \begin{tabular}[c]{@{}c@{}}71\\ (45-67)\end{tabular} &
  \begin{tabular}[c]{@{}c@{}}51\\ (45-60)\end{tabular} &
  \begin{tabular}[c]{@{}c@{}}51\\ (30-44)\end{tabular} &
  \begin{tabular}[c]{@{}c@{}} 39 \\ (34-44)\end{tabular} &
  \begin{tabular}[c]{@{}c@{}} 67 \\ (40-60)\end{tabular} &
  \begin{tabular}[c]{@{}c@{}} 45 \\ (31-44)\end{tabular} &
 
  \begin{tabular}[c]{@{}c@{}} 63 \\ (45-64)\end{tabular}&
\begin{tabular}[c]{@{}c@{}} 51 \\ (30-44)\end{tabular}
  \\ \hline
High severity &
  \begin{tabular}[c]{@{}c@{}}91\\ (44-85)\end{tabular} &
  \begin{tabular}[c]{@{}c@{}} 87\\ (45-85)\end{tabular} &
  \begin{tabular}[c]{@{}c@{}}91\\ (44-85)\end{tabular} &
  \begin{tabular}[c]{@{}c@{}}36\\ (61-85)\end{tabular} &
  \begin{tabular}[c]{@{}c@{}}87\\ (45-85)\end{tabular} &
  \begin{tabular}[c]{@{}c@{}}87\\ (45-85)\end{tabular} &
  \begin{tabular}[c]{@{}c@{}}24\\ (65-85)\end{tabular} &
  \begin{tabular}[c]{@{}c@{}}16\\ (67-85)\end{tabular} &
  \begin{tabular}[c]{@{}c@{}}36\\ (61-85)\end{tabular} &
  \begin{tabular}[c]{@{}c@{}}87 \\ (45-85)\end{tabular} &
  \begin{tabular}[c]{@{}c@{}} 87 \\ (45-85)\end{tabular} &
  \begin{tabular}[c]{@{}c@{}} 36 \\ (61-85)\end{tabular} &
  \begin{tabular}[c]{@{}c@{}} 87 \\ (45-85)\end{tabular} &
  
  \begin{tabular}[c]{@{}c@{}} 24 \\ (65-85)\end{tabular} &
  \begin{tabular}[c]{@{}c@{}} 87 \\ (45-85)\end{tabular}
  \\ \hline
\end{tabular}%
}
\end{table}

\renewcommand{\arraystretch}{1.8}
\begin{table}[htb]
\centering
\caption{PTSD Severity model performances on both text and audio modalities and their combination. Details about the specific prompts used can be found in Figure \ref{fig:ptsd-sev-p}. The underlined bold values are the best score for the specific prompt.}
\label{tab:PTSD-SEV}
\resizebox{\columnwidth}{!}{%
\begin{tabular}{cl|ccc|ccc|ccc|ccc}
\cline{3-14} 
&
  \multicolumn{1}{c}{} &
  \multicolumn{6}{c}{Intervals based on LLMs} &
  \multicolumn{6}{c}{Intervals based on \cite{10.1007/978-3-031-46933-6_21}} \\ \cline{3-14} 
 &
  \multicolumn{1}{c}{} &
  \multicolumn{3}{c}{Prompt 1} &
  \multicolumn{3}{c}{Prompt 2} &
  \multicolumn{3}{c}{Prompt 1} &
  \multicolumn{3}{c}{Prompt 2} \\
  \hline
  &\multicolumn{1}{c|}{Model} &
BA &  F1 &  MAE &  BA &  F1 &  MAE &  BA &  F1 &  MAE &  BA &  F1 &  MAE \\ 
\hline\hline
\multicolumn{1}{l}{\multirow{15}{*}{\begin{turn}{90}Text(T)\end{turn}}} &
  \multicolumn{1}{|l|}{Llama 3.1 70B} &
  62.4\% &
  0.63 &
  0.40 &
  58.1\% &
  0.56 &
  0.47 &
60\% &
\ul{\textbf{0.63}} &
\ul{\textbf{0.44}} &
  53.4\% &
  0.45 &
  0.60 \\ \cline{2-14}
 &
  \multicolumn{1}{|l|}{Gemma 2 9B} &
  50\% &
  0.56 &
  0.42 &
  48.9\% &
  0.57 &
  0.39 &
  47\% &
  0.37 &
  0.65 &
  45.7\% &
  0.33 &
  0.67 \\ \cline{2-14} 
 &
\multicolumn{1}{|l|}{GPT-4o mini} &
\ul{\textbf{69.2\%}} &
\ul{\textbf{0.73}} &
0.31 &
  68.3\% &
  0.68 &
  0.37 &
  51.2\% &
  0.60 &
  0.46 &
  54.6\% &
0.61 &
\ul{\textbf{0.44}} \\ \cline{2-14} 
 &
  \multicolumn{1}{|l|}{Mistral NeMo} &
  58.3\% &
  0.59 &
  0.45 &
  56.5\% &
  0.58 &
  0.47 &
  54.6\% &
  0.57 &
  0.50 &
  51.8\% &
  0.57 &
  0.52 \\ \cline{2-14} 
  &
  \multicolumn{1}{|l|}{Phi-3.5-MoE} &
  52.4\% &
  0.56 &
  0.50 &
  49.7\% &
  0.53 &
  0.51 &
  51.2\% &
  0.59 &
  0.49 &
  49.2\% &
  0.57 &
  0.65 \\ \cline{2-14} 
 &
  \multicolumn{1}{|l|}{Phi-3.5-mini} &
  66.9\% &
  0.60 &
  0.42 &
\ul{\textbf{69.9\%}} &
0.68 &
  0.33 &
  53.2\% &
  0.61 &
  0.47 &
  53.9\% &
  0.60 &
  0.47 \\ \cline{2-14} 
 &
  \multicolumn{1}{|l|}{Gemini 1.5 Pro} &
  57.2\% &
  0.69 &
  0.33 &
  55.6\% &
  0.63 &
  0.40 &
  45.5\% &
  0.53 &
  0.53 &
  49.7\% &
  0.51 &
  0.52 \\ \cline{2-14} 
 &
 \multicolumn{1}{|l|}{Gemini 1.5 Flash} &
  62.8\% &
  0.58 &
  0.47 &
  57\% &
  0.48 &
  0.55 &
  51.2\% &
  0.52 &
  0.53 &
  48.4\% &
  0.45 &
  0.57 \\ \cline{2-14}
 &
 \multicolumn{1}{|l|}{Deepseek\_R1} &
 57.9\% &
 0.64 &
 0.43 &
 51.8\% &
 0.59 &
 0.48 &
 52.2\% &
 0.60 &
 0.46 &
 55.1\% &
 0.62 &
 0.46 \\ \cline{2-14}
&
 \multicolumn{1}{|l|}{Gemini Flash Lite} &
 53.6\% &
 0.61 &
 0.49 &
 58.2\% &
 0.60 &
 0.49 &
 61.7\% &
 0.62 &
 0.50 &
 59.8\% &
 0.50 &
 0.65 \\ \cline{2-14}
&
 \multicolumn{1}{|l|}{Gemini 2 Flash} &
 53.2\% &
 0.58 &
 0.50 &
 51.1\% &
 0.54 &
 0.53 &
 61\% &
 0.57 &
 0.50 &
 59.4\% &
 0.54 &
 0.54 \\ \cline{2-14}
&
 \multicolumn{1}{|l|}{Gemini Flash Thinking} &
 45.7\% &
 0.52 &
 0.54 &
 51\% &
 0.55 &
 0.51 &
 49.8\% &
 0.59 &
 0.51 &
 54.2\% &
 0.59 &
 0.51 \\ \cline{2-14}
&
 \multicolumn{1}{|l|}{Gemini 2.5 Pro} &
 48.7\% &
 0.56 &
 0.50 &
 53.2\% &
 0.60 &
 0.47 &
 45.2\% &
 0.56 &
 0.53 &
 56.9\% &
 \ul{\textbf{0.64}} &
 0.53 \\ \cline{2-14} 
&
 \multicolumn{1}{|l|}{Llama 3.3 70B} &
 56.9\% &
 0.61 &
 0.46 &
 58.1\% &
 0.59 &
 0.48 &
 59\% &
 0.59 &
 0.46 &
 59.6\% &
 0.60 &
 0.44 \\ \hline \hline
\multicolumn{1}{l}{\multirow{5}{*}{\begin{turn}{90}Audio(A)\end{turn}}} &
 \multicolumn{1}{|l|}{Gemini 1.5 Pro} &
 55.2\% &
 0.72 &
 \ul{\textbf{0.28}} &
 55.1\% &
 \ul{\textbf{0.69}} &
 \ul{\textbf{0.31}} &
 40.5\% &
 0.50 &
 0.60 &
 45\% &
 0.45 &
 0.55 \\ \cline{2-14} 
&
 \multicolumn{1}{|l|}{Gemini 1.5 Flash} &
 61.4\% &
 0.56 &
 0.49 &
 60.9\% &
 0.56 &
 0.56 &
 51.3\% &
 0.56 &
 0.56 &
 53.2\% &
 0.50 &
 0.50 \\ \cline{2-14}
&
\multicolumn{1}{|l|}{Gemini Flash Lite} &
 55.1\% &
 0.61 &
 0.46 &
 48.1\% &
 0.4 &
 0.87 &
 53.8\% &
 0.56 &
 0.56 &
 53.2\% &
 0.50 &
 0.50 \\ \cline{2-14}
&
 \multicolumn{1}{|l|}{Gemini 2 Flash} &
 52.4\% &
 0.58 &
 0.48 &
 56.3\% &
 0.58 &
 0.47 &
 \textbf{\underline{67.7\%}} &
 \textbf{\underline{0.62}} &
 \textbf{\underline{0.44}} &
 51.3\% &
 0.55 &
 0.53 \\ \cline{2-14}
&
 \multicolumn{1}{|l|}{Gemini Flash Thinking} &
 50.9\% &
 0.56 &
 0.50 &
 53.1\% &
 0.57 &
 0.47 &
 55.2\% &
 0.60 &
 0.46 &
 51.1\% &
 0.55 &
 0.50 \\ \cline{2-14}
&
 \multicolumn{1}{|l|}{Gemini 2.5 Pro} &
 53.4\% &
 0.60 &
 0.46 &
 57.7\% &
 0.62 &
 0.43 &
 50.1\% &
 0.61 &
 0.46 &
 54.1\% &
 0.63 &
 0.44 \\ \hline \hline
\multicolumn{1}{l}{\multirow{5}{*}{\begin{turn}{90}A+T\end{turn}}} &
 \multicolumn{1}{|l|}{Gemini 1.5 Pro} &
 54\% &
 0.68 &
 0.35 &
 59.6\% &
 0.67 &
 0.37 &
 44\% &
 0.49 &
 0.56 &
 47.5\% &
 0.52 &
 0.53 \\ \cline{2-14} 
&
 \multicolumn{1}{|l|}{Gemini 1.5 Flash} &
 64\% &
 0.63 &
 0.41 &
 55.5\% &
 0.50 &
 0.54 &
 50.7\% &
 0.54 &
 0.50 &
 50\% &
 0.47 &
 0.55 \\ \cline{2-14}
&
 \multicolumn{1}{|l|}{Gemini 2 Flash} &
 52\% &
 0.58 &
 0.51 &
 53.2\% &
 0.50 &
 0.59 &
 49.9\% &
 0.54 &
 0.54 &
 53.8\% &
 0.52 &
 0.55 \\ \cline{2-14}
&
 \multicolumn{1}{|l|}{Gemini Flash Thinking} &
 48.9\% &
 0.50 &
 0.61 &
 52.3\% &
 0.55 &
 0.54 &
 48.4\% &
 0.54 &
 0.51 &
 51.5\% &
 0.54 &
 0.53 \\ \cline{2-14}
&
 \multicolumn{1}{|l|}{Gemini 2.5 Pro} &
 64.6\% &
 0.68 &
 0.35 &
 56.1\% &
 0.52 &
 0.57 &
 48\% &
 0.56 &
 0.51 &
 \ul{\textbf{59.8\%}} &
 0.60 &
 0.74 \\ \hline \hline
\end{tabular}%
}
\end{table}

Table~\ref{tab:PTSD-SEV} presents the results for PTSD severity classification across Text, Audio, and combined Audio+Text modalities. The performance is evaluated using Balanced Accuracy (BA), F1-score, and Mean Absolute Error (MAE) across two interval distributions: Intervals Based on LLMs and Reference Intervals. The best-performing values for each prompt are highlighted in bold and underlined.

The results demonstrate that text-based models generally outperform audio and multimodal configurations in PTSD severity classification. The best Balanced Accuracy (BA) was achieved by the Phi-3.5-mini model using Text, reaching 69.9\% with Prompt 2 under the LLM intervals. Additionally, the highest F1 score of 0.73 was attained by GPT-4o mini using Text and Prompt 1 under the same interval setting. In contrast, the lowest MAE was obtained by Gemini 1.5 Pro using Audio with Prompt 1, achieving 0.28, indicating superior precision in estimating severity levels from audio inputs.

The audio modality demonstrated strong precision for certain models, particularly with Gemini 1.5 Pro, which also maintained competitive BA and F1 scores across prompts. However, audio-based classification generally presented higher MAE compared to text-based approaches, indicating less accuracy in severity approximation.

Combining text and audio modalities did not consistently improve performance. For instance, Gemini 2.5 Pro achieved a BA of 64.6\% and an MAE of 0.35 under LLM intervals, which, while competitive, did not surpass the best text-only results. This pattern indicates that multimodal fusion does not necessarily enhance performance in PTSD severity detection.

GPT-4o mini emerged as the top performer within the text modality, particularly excelling under LLM-based intervals. Its balanced accuracy and F1 scores consistently outperformed other models, demonstrating reliable classification of PTSD severity.  
Phi-3.5-mini also demonstrated strong balanced accuracy, especially with Prompt 2, confirming its ability to effectively map PTSD severity when prompted appropriately.  
Gemini 1.5 Pro was the standout in the audio modality, achieving the lowest MAE values, indicating precise severity estimation despite being audio-based. This suggests that fine-tuning and model-specific adjustments significantly influence performance in audio-only settings.  
Gemini 2.5 Pro and other Gemini Flash models showed moderate results, often lagging behind the leading models in text-based settings but performing relatively well in audio and multimodal configurations.  
In contrast, models such as Gemma 2 9B, Mistral NeMo, and Deepseek\_R1 displayed lower accuracy and higher MAE, reflecting limited effectiveness for PTSD severity classification.  

Thinking models, such as Gemini Flash Thinking, achieved reasonable scores in some scenarios, but did not consistently outperform their non-thinking counterparts. While models like Gemini Flash Thinking showed comparable F1 scores to Gemini 2.5 Pro in audio scenarios, the added deliberative processing did not significantly enhance overall accuracy. This suggests that integrating thinking mechanisms may not always provide an advantage in PTSD severity classification, and their utility should be carefully evaluated on a case-by-case basis.


Prompt variation significantly influenced the classification results. Comparing the two prompt variants, Prompt 1 generally outperformed Prompt 2 across most metrics. 

When comparing the average performance across metrics for Prompt 1, the LLM distribution consistently demonstrated better results than the Reference distribution. The average balanced accuracy for Prompt 1 in the LLM distribution was 56.80\%, compared to 53.06\% in the Reference distribution. Similarly, the average F1 score was higher for LLM (0.60) compared to Reference (0.57), while the Mean Absolute Error (MAE) was lower for LLM (0.44) compared to Reference (0.50). 

This analysis shows that using the LLM distribution for Prompt 1 is generally more effective than relying on reference intervals. Additionally, between the two prompts, Prompt 1 shows a slight overall advantage due to its higher average scores across all three metrics, highlighting its robustness in PTSD severity classification.


The evaluation highlights that text-based models are more effective than audio-only or multimodal configurations for PTSD severity classification. GPT-4o mini and Phi-3.5-mini consistently achieved the highest balanced accuracy and F1 scores in the text modality, indicating that these models are well-suited for interpreting linguistic patterns associated with PTSD severity. Conversely, audio-based models such as Gemini 1.5 Pro demonstrated precise estimation capabilities, as reflected by their low MAE values. Multimodal combinations did not consistently outperform the best single-modality models, suggesting that optimal performance may require modality-specific fine-tuning rather than simple fusion. Additionally, the mixed performance of thinking models suggests that their utility may vary depending on the context, emphasizing the need for careful prompt and model selection in clinical applications.

\subsection{Multi-Label Classification Results}

In this task, we aimed to evaluate the models' ability to predict the presence of zero or more disorders (depression, PTSD, or both) mentioned in \ref{task3}. To achieve this, we combined the ground truth labels for the binary classification of both disorders and allowed the models to predict if the interviewee exhibited any of these conditions. Once the predictions were made, we calculated the balanced accuracy (BA) and F1 score for each disorder separately, based on the predicted labels for both depression and PTSD.

We also calculated the Balanced Accuracy (BA) and F1 score when treating the problem as a multiclass classification task with four classes: Depression, PTSD, both, or None. Additionally, for the multi-label classification task, we calculated BA and F1 scores based on partial correctness. For example, if the accurate labels were Depression and PTSD, but the model predicted only PTSD, it was credited as 50\% correct since it partially matched the ground truth. This approach allows us to evaluate performance across different classification settings.

This framework allowed us to compare the models' ability to handle complex cases where more than one condition might be present, providing valuable insights into their predictive accuracy for mental health diagnostics.

\renewcommand{\arraystretch}{1.8}
\begin{table}[htb]
\centering
\caption{Multi label model performance on both Text and Audio modalities and their combination. Details about the specific prompts used can be found in Figure \ref{fig:multi-p}. The underlined bold values are the best score for the specific column.}
\label{tab:Multi}
\resizebox{\columnwidth}{!}{%
\begin{tabular}{llcccc|cccc|cccc|cccc}
\cline{3-18}
 && \multicolumn{4}{c}{Depression} & \multicolumn{4}{c}{PTSD} & \multicolumn{4}{c}{Multiclass} & \multicolumn{4}{c}{Multi-Label} \\
\cline{3-18}
 & & \multicolumn{2}{c}{Prompt 1} & \multicolumn{2}{c}{Prompt 2} & \multicolumn{2}{c}{Prompt 1} & \multicolumn{2}{c}{Prompt 2} & \multicolumn{2}{c}{Prompt 1} & \multicolumn{2}{c}{Prompt 2} & \multicolumn{2}{c}{Prompt 1} & \multicolumn{2}{c}{Prompt 2} \\
\hline
 & \multicolumn{1}{c|}{Model} & BA & F1 & BA & F1 & BA & F1 & BA & F1 & BA & F1 & BA & F1 & BA & F1 & BA & F1 \\
\hline \hline
\multirow{15}{*}{\begin{turn}{90}Text (T)\end{turn}} &
  \multicolumn{1}{l|}{Llama 3.1 70B} & 69.9\% & 0.59 & 72.3\% & 0.65 & \textbf{\ul{74\%}} & \textbf{\ul{0.75}} & 68.1\% & 0.73 & 43.2\% & 0.53 & 40.2\% & 0.55 & \textbf{\ul{72\%}} & 0.83 & \textbf{\ul{73\%}} & 0.74 \\ \cline{2-18}
 & \multicolumn{1}{l|}{Gemma 2 9B} & 53.2\% & 0.28 & 64.9\% & 0.51 & 60.6\% & 0.45 & 72.3\% & 0.75 & 29.5\% & 0.22 & 37.9\% & 0.45 & 57\% & 0.88 & 68\% & \textbf{\underline{0.80}} \\ \cline{2-18}
 & \multicolumn{1}{l|}{GPT-4o mini} & 68.7\% & 0.56 & 72.5\% & 0.67 & 66.9\% & 0.67 & \textbf{\underline{74.2\%}} & \textbf{\underline{0.77}} & 39.9\% & 0.49 & 45\% & 0.60 & 68\% & 0.82 & \textbf{\ul{73\%}} & 0.77 \\ \cline{2-18}
 & \multicolumn{1}{l|}{Mistral NeMo} & 67.2\% & 0.59 & 69.4\% & 0.67 & 70.4\% & 0.75 & 63.9\% & 0.72 & 37.7\% & 0.51 & 38.7\% & 0.55 & 68.8\% & 0.75 & 66.5\% & 0.63 \\ \cline{2-18}
 & \multicolumn{1}{l|}{Phi-3.5-MoE} & 66.9\% & 0.56 & 68.6\% & 0.60 & 66.8\% & 0.70 & 62.4\% & 0.68 & 38.2\% & 0.48 & 37.9\% & 0.50 & 67\% & 0.77 & 65\% & 0.70 \\ \cline{2-18}
 & \multicolumn{1}{l|}{Phi-3.5-mini} & 52.3\% & 0.21 & 64.2\% & 0.52 & 52.6\% & 0.22 & 66.4\% & 0.66 & 26.3\% & 0.15 & 35.8\% & 0.41 & 53\% & \textbf{\underline{0.91}} & 65\% & \textbf{\underline{0.80}} \\ \cline{2-18}
 & \multicolumn{1}{l|}{Gemini 1.5 Pro} & 66.1\% & 0.58 & 72.3\% & \textbf{\underline{0.74}} & 67.6\% & 0.70 & 66.7\% & 0.73 & 41.2\% & 0.45 & 45\% & 0.62 & 67\% & 0.77 & 69\% & 0.60 \\ \cline{2-18}
 & \multicolumn{1}{l|}{Gemini 1.5 Flash} & 64.4\% & 0.50 & 74.3\% & 0.70 & 72\% & 0.75 & 69.1\% & 0.75 & 41.2\% & 0.45 & 45.4\% & 0.58 & 68\% & 0.81 & 72\% & 0.73 \\ \cline{2-18}
 & \multicolumn{1}{l|}{Deepseek\_R1} & 70.2\% & 0.60 & 74.5\% & 0.70 & 64.9\% & 0.70 & 71.8\% & 0.76 & 43.3\% & 0.50 & \textbf{\underline{47.7\%}} & \textbf{\underline{0.63}} & 67\% & 0.77 & \textbf{\ul{73\%}} & 0.75 \\ \cline{2-18}
 & \multicolumn{1}{l|}{Gemini Flash Lite} & 69.6\% & 0.60 & 70.7\% & 0.62 & 70\% & 0.72 & 68\% & 0.70 & 42.9\% & 0.52 & 43.7\% & 0.52 & 70\% & 0.81 & 69\% & \textbf{\underline{0.80}} \\ \cline{2-18}
 & \multicolumn{1}{l|}{Gemini 2 Flash} & 66.7\% & 0.59 & 68.8\% & 0.71 & 66.9\% & 0.65 & 68.3\% & 0.70 & 38.1\% & 0.49 & 40.5\% & 0.56 & 67\% & 0.81 & 69\% & 0.70 \\ \cline{2-18}
 & \multicolumn{1}{l|}{Gemini Flash Thinking} & 73.4\% & \textbf{\underline{0.72}} & 69.7\% & 0.64 & 64.1\% & 0.69 & 67\% & 0.70 & 39.5\% & 0.55 & 40.3\% & 0.52 & 69\% & 0.74 & 68\% & 0.75 \\ \cline{2-18}
 & \multicolumn{1}{l|}{Gemini 2.5 Pro} & \textbf{\underline{74.8\%}} & 0.71 & 71.8\% & 0.65 & 67\% & 0.72 & 70.7\% & 0.74 & 44.2\% & 0.59 & 43.4\% & 0.54 & 71\% & 0.73 & 71\% & 0.78 \\ \cline{2-18}
 & \multicolumn{1}{l|}{Llama 3.3 70B} & 69\% & 0.59 & \textbf{\underline{75.6\%}} & 0.69 & 69.8\% & 0.73 & 69.5\% & 0.73 & 41.3\% & 0.51 & 44.1\% & 0.59 & 69\% & 0.80 & \textbf{\ul{73\%}} & 0.78 \\ \hline \hline

\multirow{6}{*}{\begin{turn}{90}Audio (A)\end{turn}} &
  \multicolumn{1}{l|}{Gemini 1.5 Pro} & 71.9\% & 0.72 & 67.3\% & 0.70 & 67.4\% & 0.73 & 67.8\% & 0.74 & 42.5\% & 0.58 & 39.2\% & 0.57 & 70\% & 0.67 & 68\% & 0.60 \\ \cline{2-18}
 & \multicolumn{1}{l|}{Gemini 1.5 Flash} & 71.7\% & 0.70 & 71.2\% & 0.69 & 68.1\% & 0.73 & 63.1\% & 0.67 & 44.8\% & 0.57 & 42.4\% & 0.52 & 70\% & 0.71 & 67\% & 0.72 \\ \cline{2-18}
 & \multicolumn{1}{l|}{Gemini Flash Lite} & 70.7\% & 0.69 & 69.3\% & 0.71 & 71.1\% & 0.71 & 71.7\% & 0.75 & 40.4\% & 0.59 & 39\% & 0.61 & 71\% & 0.72 & 71\% & 0.63 \\ \cline{2-18}
 & \multicolumn{1}{l|}{Gemini 2 Flash} & 70.7\% & 0.67 & 71.3\% & 0.71 & 67.3\% & 0.71 & 69.7\% & 0.72 & 40.5\% & 0.56 & 45.8\% & 0.60 & 69\% & 0.73 & 71\% & 0.74 \\ \cline{2-18}
 & \multicolumn{1}{l|}{Gemini Flash Thinking} & 73.4\% & 0.72 & 71.6\% & 0.71 & 69.1\% & 0.75 & 69.2\% & 0.74 & 41.9\% & \textbf{\underline{0.61}} & 44.5\% & 0.61 & 71\% & 0.68 & 70\% & 0.68 \\ \cline{2-18}
 & \multicolumn{1}{l|}{Gemini 2.5 Pro} & 72\% & 0.67 & 74.1\% & 0.69 & 69.7\% & 0.74 & 68.6\% & 0.72 & \textbf{\underline{47.9\%}} & 0.58 & 45.6\% & 0.58 & 71\% & 0.76 & 71\% & 0.76 \\ \hline \hline

\multirow{5}{*}{\begin{turn}{90}A + T\end{turn}} &
  \multicolumn{1}{l|}{Gemini 1.5 Pro} & 72.3\% & 0.71 & 74\% & \textbf{\underline{0.74}} & 67.1\% & 0.72 & 69.8\% & 0.72 & 43\% & 0.56 & 45.9\% & 0.52 & 70\% & 0.72 & 72\% & 0.68 \\ \cline{2-18}
 & \multicolumn{1}{l|}{Gemini 1.5 Flash} & 72.4\% & 0.64 & 72\% & 0.69 & 70.6\% & 0.74 & 65.2\% & 0.70 & 46.5\% & 0.56 & 44.1\% & 0.55 & \textbf{\underline{72\%}} & 0.79 & 69\% & 0.73 \\ \cline{2-18}
 & \multicolumn{1}{l|}{Gemini 2 Flash} & 72.6\% & 0.69 & 72.1\% & 0.70 & 67.7\% & 0.70 & 64.1\% & 0.67 & 44.8\% & 0.56 & 41.7\% & 0.55 & 70\% & 0.77 & 68\% & 0.73 \\ \cline{2-18}
 & \multicolumn{1}{l|}{Gemini Flash Thinking} & 73.8\% & 0.68 & 72.6\% & 0.67 & 66.7\% & 0.71 & 69.3\% & 0.74 & 43.1\% & 0.55 & 45\% & 0.57 & 70\% & 0.76 & 71\% & 0.76 \\ \cline{2-18}
 & \multicolumn{1}{l|}{Gemini 2.5 Pro} & 70.9\% & 0.65 & 72.9\% & 0.66 & 69.9\% & 0.74 & 67.9\% & 0.73 & 43\% & 0.55 & 45.7\% & 0.56 & 70\% & 0.76 & 70\% & 0.77 \\ \hline
\end{tabular}%
}
\end{table}

In the depression detection task summarized in Table \ref{tab:Multi}, Llama 3.3 70B achieved the best results in the text modality with a balanced accuracy (BA) of 75.6\% and an F1 score of 0.69 on Prompt 2. For PTSD detection, GPT-4o mini led with a BA of 74.2\% and an F1 score of 0.77 on Prompt 2. In the multiclass and multi-label tasks, Gemini 2.5 Pro achieved the highest BA of 47.9\% in the multiclass task while, Phi-3.5-mini achieved an F1 score of 0.91 in the multi-label task.

Gemini 2.5 Pro emerged as the most consistent model when applying the text modality, consistently achieving high scores across multiple tasks and prompts, validating its reliability in this setup.

When both modalities are combined, Gemini 1.5 Flash again shows enhanced performance, achieving a BA of 72\% on Prompt 1 and an F1 score of 0.79 on the multi-label task under this configuration. This multimodal approach, which leverages both textual and vocal data, appears to provide a more comprehensive analysis, potentially increasing the accuracy and reliability of mental health assessments.

Overall, models in the text modality generally performed better in PTSD detection, while audio-based models showed better performance in depression detection. However, the combined modalities often outperformed individual text or audio inputs, suggesting that integrating these approaches may offer the most effective means for mental health diagnostics.

Among the evaluated models, Gemini 1.5 Flash consistently showed the strongest results in both single and multi-label tasks, particularly excelling in the text modality. GPT-4o mini also demonstrated high performance, especially in PTSD detection with Prompt 2. Deepseek\_R1 presented competitive results in both multiclass and multi-label settings, especially when combining text and audio data.

Regarding the comparison between thinking and non-thinking models, the "thinking" models, such as Gemini Flash Thinking, did not consistently outperform their non-thinking counterparts. Although they occasionally achieved competitive F1 scores, like 0.72 in the audio modality, their performance varied across tasks and prompts. On the other hand, non-thinking models, especially Gemini 1.5 Pro, often provided more stable and reliable results. This suggests that the added complexity in thinking models may not consistently enhance accuracy, particularly in multi-label mental health diagnostics.


The influence of prompt design was also evident in performance variations. On average, Prompt 2 produced slightly better F1 scores across most models and tasks, particularly for PTSD and multi-label classification. However, Prompt 1 tended to yield higher balanced accuracy scores in the depression detection subtask. Prompt two may help models better identify relevant disorder cues for precision (F1), whereas Prompt 1 encourages broader recall patterns beneficial for class balance. Overall, Prompt 2 showed slightly stronger general performance, especially in complex multi-condition settings.


The analysis highlights the effectiveness of integrating text and audio modalities for mental health assessment, as combined approaches generally yielded superior results compared to single modalities. Gemini 1.5 Flash emerged as the most reliable model across diverse tasks, demonstrating consistent performance in multi-label classification scenarios. While thinking models offered some advantages, they did not consistently outperform non-thinking variants, indicating that their application should be considered based on specific diagnostic requirements. These findings underscore the importance of careful model selection and the potential benefits of multimodal approaches in complex mental health diagnostics.

\subsection{Disagreement}

\begin{figure}[htb]
\centering

\begin{subfigure}[b]{0.6\textwidth}
    \centering
    \includegraphics[width=\textwidth]{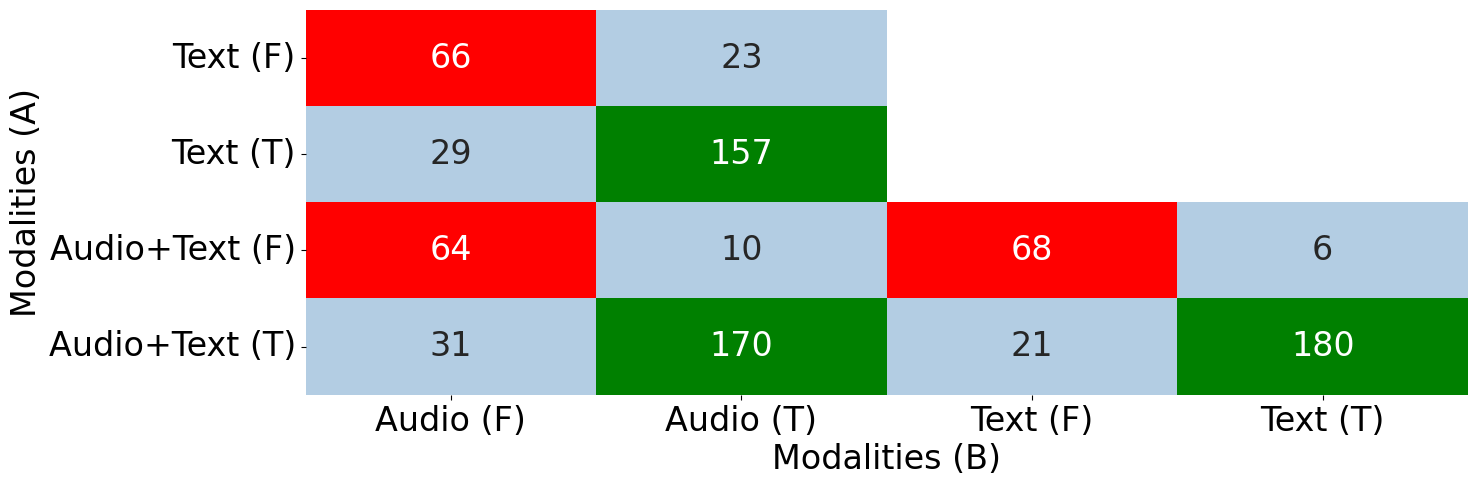}
    \caption{}
    \label{fig:good-co}
\end{subfigure}%
\hfill
\begin{subfigure}[b]{0.4\textwidth}
    \centering
    \includegraphics[width=\textwidth]{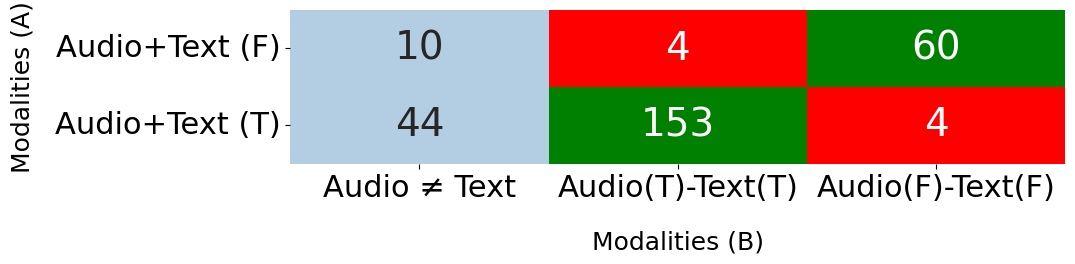}
    \caption{}
    \label{fig:good-at}
\end{subfigure}
\caption{(a) Co-occurrence matrix illustrating where the Audio and Text modalities and their combination independently predict binary depression outcomes correctly or incorrectly under zero-shot inference conditions with the Gemini 2.5 Pro model using Prompt 1.
(b) Visualization of the combined (Audio + Text) modality’s predictions relative to the individual Audio and Text modalities, highlighting scenarios of agreement, disagreement, and how the combined modality addresses these differences under the same conditions.}
\label{fig:good-ex}
\end{figure}

\begin{figure}[htb]
\centering
\begin{subfigure}[b]{0.6\textwidth}
    \centering    \includegraphics[width=\textwidth]{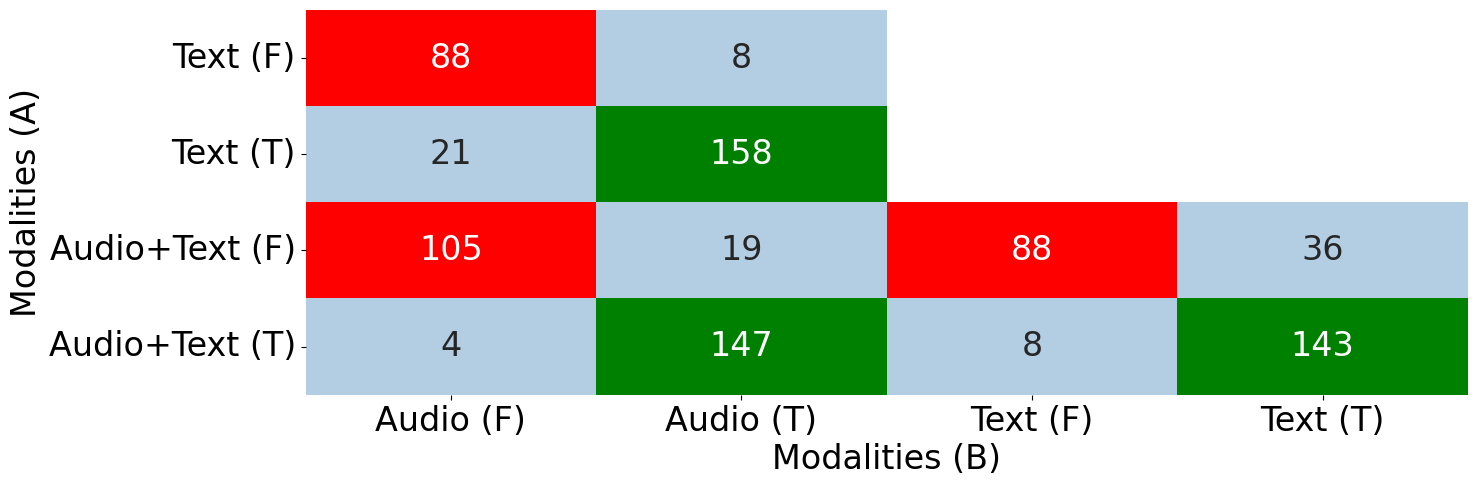}
    \caption{}
   \label{fig:bad-co}
\end{subfigure}
\hfill
\begin{subfigure}[b]{0.38\textwidth}
    \centering \includegraphics[width=\textwidth]{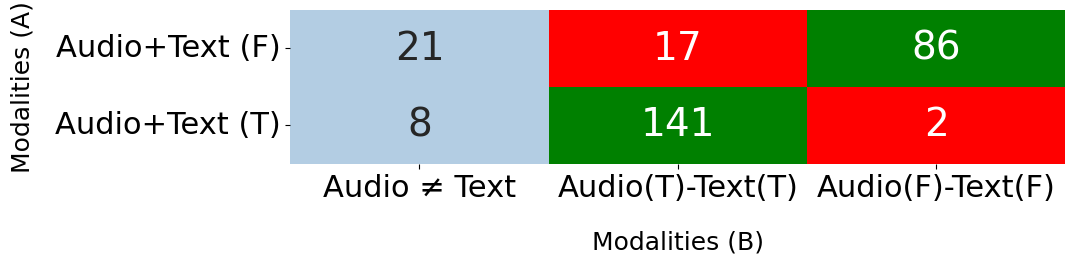}    
    \caption{}
    \label{fig:bad-at}
\end{subfigure}
\caption{(a) Co-occurrence matrix illustrating where the Audio and Text modalities and their combination independently predict binary depression outcomes correctly or incorrectly under zero-shot inference conditions with the Gemini 1.5 Flash model using Prompt 2.
(b) Visualization of the combined (Audio + Text) modality’s predictions relative to the individual Audio and Text modalities, highlighting scenarios of agreement, disagreement, and how the combined modality addresses these differences under the same conditions.}
\label{fig:bad-ex}
\end{figure}

The co-occurrence matrix in Figures \ref{fig:good-co} and \ref{fig:bad-co} panels visualizes the instances of correct and incorrect predictions by each modality for binary depression classification using the Gemini 1.5 Flash model on Prompt 3 and 2, respectively. The colors represent different outcomes of these predictions:
Red colored cells indicate instances where both modalities incorrectly predicted the sample.
Green colored cells highlight cases where both modalities correctly identified the sample.
Blue colored cells denote instances where one modality outperformed the other by correctly predicting a sample that the other modality missed.

Figures \ref{fig:good-at} and  \ref{fig:bad-at} present the analysis of the Audio+Text modality's performance in resolving disagreements between the individual Audio and Text modalities. The colors in the heatmap signify different outcomes of predictions across these modalities:
Green colored cells indicate instances where all three modalities (Audio, Text, and Audio+Text) either correctly or incorrectly predicted the sample.
Red colored cells show the number of samples where the combined modality predicted differently than both the Audio and Text modalities when they were either correct or incorrect.
Blue colored cells highlight cases where the combined modality resolved the disagreement between the Audio and Text modalities, either by correctly predicting what one modality missed or incorrectly predicting what one modality got right.

\

For figure \ref{fig:good-co}, the Modal Superiority Score (MSS) was applied to evaluate the performance differences among the modalities. The MSS metric, as detailed in Equation \ref{eq:MSS}, measures the effectiveness of one modality over another when there is disagreement. The findings indicated that, compared to the separate modalities, the Audio+Text combined modality performed better. The advantages of modality integration were highlighted by MSS values, which demonstrated a 52.5\% superiority of Audio+Text over Audio alone and a 55.5\% superiority over Text alone. Text displayed a slight advantage over Audio, with an MSS value of 11.5\%, indicating that Text was marginally more successful at accurately predicting results than Audio.


To further illustrate how the combined modality improved performance, we applied the Disagreement Resolution Score (DRS) and the Modal Superiority Score (MSS) metrics on figure \ref{fig:good-at}. The DRS, detailed in Equation \ref{eq:DRS}, explicitly addresses the combined modality's ability to resolve conflicts between the two other modalities. It calculated a significant positive value of 62.96\%, indicating a slight net benefit in the combined modality's resolution of disagreements. Suggests that the combined Audio+Text modality was able to correctly resolve over half of the instances where Audio and Text disagreed.

Additionally, the MSS between the combined modality and the joint prediction agreement of the Audio and Text modalities showed a substantial positive score of 43.7\%. This result underscores a significant enhancement in performance by the combined modality over the consensus of the individual modalities, reflecting a superior capability to harness the strengths of both modalities effectively.

In the assessment of modality performance using the Modal Superiority Score (MSS) for figure \ref{fig:bad-ex}, we quantified the efficacy of individual and combined modalities. The findings are summarized as follows:
 
The MSS value calculated for Text against Audio was 44.83\%. This positive value demonstrates that Text outperformed Audio. The MSS values indicate that the performance declined when Audio and Text were combined, yielding -65.22\% against Audio alone and -63.64\% against Text alone. These negative scores highlight a decrease in performance when the modalities are combined, indicating that the integration of Text with Audio may have introduced complexities that reduced the effectiveness of the prediction capability.  

Using the Disagreement Resolution Score (DRS), we evaluated the combined Audio+Text modality's ability to resolve conflicts between the Audio and Text modalities, resulting in a significant negative score of approximately -44.83\%. This significant negative score indicates that integrating Audio and Text modalities in this experiment negatively impacted the model's ability to resolve disagreements. This finding underscores the complexities and potential challenges associated with modality integration, demonstrating that combining different modalities does not always enhance predictive accuracy.

\subsection{Few-shot Prompts Results}

Tables \ref{tab:FS-score-merged}  present the few-shot (FS) performance scores for various tasks using specifically designed prompts, as described in Section \ref{fs-p}. These evaluations were conducted on the full dataset of 275 samples, including those used in the prompts, due to the models’ tendency to struggle with them in zero-shot (ZS) settings. FS scores are shown alongside the change from the ZS baseline. The Gemini Flash Lite (Audio) model was excluded from these evaluations due to high variance and inconsistency in its ZS performance, which made comparative analysis unreliable.

In the binary depression task, Gemini 2.5 Pro (Audio) achieved the best result with 81.1\% balanced accuracy and 0.81 F1 on Prompt 3. The most significant improvement was from Gemini Flash 1.5 (Text) on Prompt 2, with +4.7\% accuracy and +0.05 F1, while Phi-3.5-MoE (Text) showed a slight decline. Audio had the highest peak, but text models were more consistent.

For binary PTSD, Gemini 2.5 Pro (Text) led with 79.9\% accuracy and 0.79 F1 on Prompt 1. Gemini Flash Lite (Text) had the most substantial improvement on Prompt 2 (+11.2\%, +0.07 F1). Text-based models outperformed audio, both in final scores and gains.

In depression severity, the top score came from Phi-3.5-MoE (Text) on Prompt 2 (42.9\%, 0.46 F1), while Gemini Flash 1.5 (Text) improved the most on Prompt 1. GPT-4o mini had a notable drop on Prompt 2. Text remained the more stable and effective modality in this task.

For PTSD severity, Gemini 2.5 Pro (Text) on Prompt 1 reached 65\% accuracy and 0.70 F1 (LLM), with strong alignment in reference distribution. It also showed the highest gain overall (+16.3\%, +1.4 F1). Text models consistently outperformed audio here as well.

In the multi-label task, Gemini 2.5 Pro (Text) led with 74\% accuracy and 0.81 F1 on Prompt 2, while Phi-3.5-MoE (Text) recorded the highest F1 gain (+0.14). Minor performance drops were seen in Gemini Flash Lite (Text). Text dominated this task, with multiple models surpassing 0.80 F1.

In summary, Gemini 2.5 Pro—particularly its text and audio variants—delivered the most consistent and high-performing results across tasks. Prompt 1 was most effective for PTSD binary and severity, while Prompt 3 yielded the best results for depression binary, and Prompt 2 was strongest in multi-label and severity tasks. Textual inputs generally offered broader and more stable improvements across functions, while audio demonstrated isolated peak performance, especially in depression classification. These findings demonstrate the effectiveness of tailored few-shot prompts and reinforce the potential of prompt-conditioned LLMs for scalable mental health assessment.

\subsection{Comparative Evaluation}

In this section, we undertake a comparative analysis of our model’s performance against other established methodologies, as detailed in tables \ref{tab:d-papers}. A thorough literature search revealed a scarcity of studies that evaluated models across the entire E-DAIC or DAIC-WOZ datasets. To facilitate a robust comparison, we selected our best-performing model and the corresponding prompt. We evaluated this model in the development sets of both the E-DAIC, which contains 56 samples, and DAIC-WOZ, which consists of 35 samples. This approach allows us to directly compare our results with those obtained from other models tested under similar conditions. For our evaluation metric, we used the binary score F1, considering the task's focus on binary depression
classification, a prevalent method in existing research. We prioritized scores that were consistently high across both datasets to ensure a balanced evaluation, avoiding instances where a model performed exceptionally well on one dataset but poorly on another. This method provides a clear benchmark to assess the relative effectiveness of our proposed approaches. 

Table \ref{tab:ptsd-c} offers a detailed comparative view of PTSD diagnosis performance across various methodologies, as shown below. Our zero-shot approach with GPT-4o mini is not only competitive but also surpasses the results reported by other significant studies. Notably, the CALLM model developed by Wu et al., which utilizes fine-tuning, shows a slight improvement over our approach with a balanced accuracy of 77\% and an F1 score of 0.70. However, our zero-shot model still demonstrates a robust capability comparable to a finely tuned system like CALLM, outperforming other established benchmarks in the field. Underscores the potential of non-fine-tuned models to achieve high effectiveness, highlighting a significant advance in using LLMs for PTSD classification without extensive training on specific datasets.

\begin{landscape}
\begin{table}[htb]
\centering
\Large
\caption{FS model performance evaluation. Values are: FS score (change from ZS score). The best score for each prompt is highlighted in bold and underlined.}
\label{tab:FS-score-merged}
\renewcommand{\arraystretch}{1.8}
\newcolumntype{P}[1]{>{\centering\arraybackslash}p{#1}}
\resizebox{\columnwidth}{!}{
\begin{tabular}{|c|c|cc|cc|cc|cc|cc|cc|cc|}
\hline
\multicolumn{2}{|c|}{\textbf{Task \& Prompt}} &
\multicolumn{2}{c|}{\shortstack{\textbf{Gemini Flash 1.5} \\ \textbf{(Text)}}} &
\multicolumn{2}{c|}{\shortstack{\textbf{GPT-4o mini} \\ \textbf{(Text)}}} &
\multicolumn{2}{c|}{\shortstack{\textbf{Phi-3.5-MoE} \\ \textbf{(Text)}}} &
\multicolumn{2}{c|}{\shortstack{\textbf{Gemini Flash 1.5} \\ \textbf{(Audio)}}} &
\multicolumn{2}{c|}{\shortstack{\textbf{Gemini 2.5 Pro} \\ \textbf{(Text)}}} &
\multicolumn{2}{c|}{\shortstack{\textbf{Gemini Flash Lite} \\ \textbf{(Text)}}} &
\multicolumn{2}{c|}{\shortstack{\textbf{Gemini 2.5 Pro} \\ \textbf{(Audio)}}} \\
\hline

\multirow{3}{*}{Depression Binary} 
& P1 & \shortstack{76.1\% \\ (+2.1\%)} & \shortstack{0.65 \\ (+0.02)} & 
\shortstack{71.4\% \\ (+6.6\%)} & \shortstack{0.60 \\ (+0.10)} & 
\shortstack{71.0\% \\ (-3.5\%)} & \shortstack{0.60 \\ (-0.04)} & 
\shortstack{76.0\% \\ (+1.1\%)} & \shortstack{0.66 \\ (0)} & 
\shortstack{\textbf{\ul{78.47\%}} \\ (+3.6\%)} & \shortstack{0.68 \\ (0)} & 
\shortstack{76.3\% \\ (-0.6\%)} & \shortstack{\textbf{\ul{0.74}} \\ (+0.01)} & 
\shortstack{75.1\% \\ (-0.04\%)} & \shortstack{0.73 \\ (+0.03)} \\ \cline{2-16}

& P2 & \shortstack{\textbf{\ul{77.4\%}} \\ (+4.7\%)} & \shortstack{\textbf{\ul{0.67}} \\ (+0.05)} & 
\shortstack{65.5\% \\ (-9\%)} & \shortstack{0.56 \\ (-0.07)} & 
\shortstack{74.7\% \\ (+1.7\%)} & \shortstack{0.64 \\ (+0.02)} & 
\shortstack{71.4\% \\ (+0.9\%)} & \shortstack{0.60 \\ (0)} & 
\shortstack{74.1\% \\ (+1.1\%)} & \shortstack{0.66 \\ (+0.01)} & 
\shortstack{71.5\% \\ (+2.4\%)} & \shortstack{0.63 \\ (+0.05)} & 
\shortstack{75.9\% \\ (+0.06\%)} & \shortstack{0.65 \\ (+0.05)} \\ \cline{2-16}

& P3 & \shortstack{77.7\% \\ (+4.1\%)} & \shortstack{0.68 \\ (+0.05)} & 
\shortstack{74.2\% \\ (+9\%)} & \shortstack{0.64 \\ (+0.14)} & 
\shortstack{76.9\% \\ (+0.9\%)} & \shortstack{0.67 \\ (0)} & 
\shortstack{75.4\% \\ (+0.7\%)} & \shortstack{0.65 \\ (0)} & 
\shortstack{79.0\% \\ (+1.75\%)} & \shortstack{0.78 \\ (+0.01)} & 
\shortstack{77.7\% \\ (+0.4\%)} & \shortstack{0.77 \\ (+0.02)} & 
\shortstack{\textbf{\ul{81.1\%}} \\ (+2.2\%)} & \shortstack{\textbf{\ul{0.81}} \\ (+0.05)} \\ \hline

\multirow{2}{*}{PTSD Binary} 
& P1 & \shortstack{\textbf{\ul{80.0\%}} \\ (+11\%)} & \shortstack{0.71 \\ (+0.13)} & 
\shortstack{78.2\% \\ (+1.6\%)} & \shortstack{0.70 \\ (+0.03)} & 
\shortstack{76.8\% \\ (+5.6\%)} & \shortstack{0.68 \\ (-0.08)} & 
\shortstack{75.3\% \\ (+2.4\%)} & \shortstack{0.65 \\ (+0.03)} & 
\shortstack{79.9\% \\ (+10.5\%)} & \shortstack{\textbf{\ul{0.79}} \\ (+0.04)} & 
\shortstack{67.7\% \\ (+2\%)} & \shortstack{0.56 \\ (-0.09)} & 
\shortstack{74.6\% \\ (+5.9\%)} & \shortstack{0.61 \\ (-0.08)} \\ \cline{2-16}

& P2 & \shortstack{77.5\% \\ (+7\%)} & \shortstack{0.68 \\ (+0.70)} & 
\shortstack{76.0\% \\ (-1\%)} & \shortstack{0.67 \\ (-0.01)} & 
\shortstack{72.0\% \\ (+2.8\%)} & \shortstack{0.61 \\ (+0.05)} & 
\shortstack{71.9\% \\ (+4.7\%)} & \shortstack{0.62 \\ (+0.05)} & 
\shortstack{\textbf{\ul{76.9\%}} \\ (+4\%)} & \shortstack{\textbf{\ul{0.78}} \\ (+0.02)} & 
\shortstack{76.7\% \\ (+11.2\%)} & \shortstack{0.73 \\ (+0.07)} & 
\shortstack{75.2\% \\ (+5\%)} & \shortstack{0.77 \\ (+0.01)} \\ \hline

\multirow{2}{*}{Depression Severity} 
& P1 & \shortstack{38.4\% \\ (+3.1\%)} & \shortstack{0.42 \\ (+0.07)} & 
\shortstack{34.0\% \\ (-10\%)} & \shortstack{0.32 \\ (-0.18)} & 
\shortstack{38.9\% \\ (-10\%)} & \shortstack{0.33 \\ (-0.16)} & 
\shortstack{34.0\% \\ (-1\%)} & \shortstack{0.42 \\ (0)} & 
\shortstack{44.2\% \\ (+5.2\%)} & \shortstack{0.52 \\ (+0.06)} & 
\shortstack{43.6\% \\ (+5.4\%)} & \shortstack{0.43 \\ (+0.03)} & 
\shortstack{\textbf{\ul{47.3\%}} \\ (+6.1\%)} & \shortstack{\textbf{\ul{0.53}} \\ (+0.01)} \\ \cline{2-16}

& P2 & \shortstack{34.7\% \\ (0\%)} & \shortstack{0.44 \\ (+0.14)} & 
\shortstack{28.1\% \\ (-13.6\%)} & \shortstack{0.12 \\ (-0.38)} & 
\shortstack{42.9\% \\ (+10\%)} & \shortstack{0.46 \\ (+0.04)} & 
\shortstack{41.8\% \\ (+6.6\%)} & \shortstack{0.41 \\ (+0.06)} & 
\shortstack{\textbf{\ul{48.1\%}} \\ (+2.9\%)} & \shortstack{\textbf{\ul{0.54}} \\ (-0.02)} & 
\shortstack{37.5\% \\ (+3.2\%)} & \shortstack{0.40 \\ (+0.11)} & 
\shortstack{41.6\% \\ (-2\%)} & \shortstack{0.51 \\ (+0.05)} \\ \hline

\multirow{2}{*}{PTSD Severity} 
& P1 & 
\shortstack{LLM: \\ 61.2\% \\ (-1.6\%)} & 
\shortstack{LLM: \\ 0.51 \\ (-0.07)} & 
\shortstack{LLM: \\ 53.4\% \\ (-15.8\%)} & 
\shortstack{LLM: \\ 0.30 \\ (-0.29)} & 
\shortstack{LLM: \\ 64.7\% \\ (+11.5\%)} & 
\shortstack{LLM: \\ 0.62 \\ (+0.23)} & 
\shortstack{LLM: \\ 64.6\% \\ (+3.2\%)} & 
\shortstack{\textbf{\ul{LLM:}} \\ \textbf{\ul{0.70}} \\ (+0.14)} & 
\shortstack{\textbf{\ul{LLM:}} \\ \textbf{\ul{65\%}} \\ (+16.3\%)} & 
\shortstack{\textbf{\ul{LLM:}} \\ \textbf{\ul{0.70}} \\ (+1.4)} & 
\shortstack{LLM: \\ 57.3\% \\ (+4.1)} & 
\shortstack{LLM: \\ 0.59 \\ (-0.01)} & 
\shortstack{LLM: \\ 62.2\% \\ (+8.8\%)} & 
\shortstack{\textbf{\ul{LLM:}} \\ \textbf{\ul{0.70}} \\ (+0.10)} \\

&     & 
\shortstack{REF: \\ 49.8\% \\ (-1.4\%)} & 
\shortstack{REF: \\ 0.53 \\ (0)} & 
\shortstack{REF: \\ 53.5\% \\ (+2.3\%)} & 
\shortstack{REF: \\ 0.50 \\ (-0.10)} & 
\shortstack{REF: \\ 54.4\% \\ (+2.7\%)} & 
\shortstack{REF: \\ 0.49 \\ (-0.11)} & 
\shortstack{REF: \\ 47.8\% \\ (-3.5\%)} & 
\shortstack{REF: \\ 0.54 \\ (+0.06)} & 
\shortstack{\textbf{\ul{REF:}} \\ \textbf{\ul{58.4\%}} \\ (+13.2\%)} & 
\shortstack{\textbf{\ul{REF:}} \\ \textbf{\ul{0.62}} \\ (+0.06)} & 
\shortstack{\textbf{\ul{REF:}} \\ \textbf{\ul{58.4\%}} \\ (+1.5)} & 
\shortstack{REF: \\ 0.60 \\ (-0.04)} & 
\shortstack{REF: \\ 56.5\% \\ (+6.4\%)} & 
\shortstack{REF: \\ 0.62 \\ (+0.01)} \\ \cline{2-16}

& P2 & 
\shortstack{LLM: \\ 55.4\% \\ (-1.6\%)} & 
\shortstack{LLM: \\ 0.58 \\ (+0.10)} & 
\shortstack{LLM: \\ 51.1\% \\ (-17.2\%)} & 
\shortstack{LLM: \\ 0.25 \\ (-0.43)} & 
\shortstack{LLM: \\ 59.9\% \\ (+11.3\%)} & 
\shortstack{LLM: \\ 0.57 \\ (+0.21)} & 
\shortstack{\textbf{\ul{LLM:}} \\ \textbf{\ul{65\%}} \\ (+4.1\%)} & 
\shortstack{LLM: \\ 0.62 \\ (+0.15)} & 
\shortstack{LLM: \\ 60\% \\ (+6.8\%)} & 
\shortstack{\textbf{\ul{LLM:}} \\ \textbf{\ul{0.64}} \\ (+0.04)} & 
\shortstack{LLM: \\ 55.1\% \\ (-3.1)} & 
\shortstack{LLM: \\ 0.54 \\ (-0.06)} & 
\shortstack{LLM: \\ 55.1\% \\ (-3.1\%)} & 
\shortstack{LLM: \\ 0.63 \\ (+0.01)} \\

&     & 
\shortstack{REF: \\ 47.8\% \\ (-0.6\%)} & 
\shortstack{REF: \\ 0.48 \\ (+0.03)} & 
\shortstack{REF: \\ 46.6\% \\ (-8\%)} & 
\shortstack{REF: \\ 0.36 \\ (-0.25)} & 
\shortstack{REF: \\ 57.7\% \\ (+8.9\%)} & 
\shortstack{REF: \\ 0.59 \\ (+0.02)} & 
\shortstack{REF: \\ 54.7\% \\ (+1.5\%)} & 
\shortstack{REF: \\ 0.58 \\ (+0.04)} & 
\shortstack{REF: \\ 54\% \\ (-2.9)} & 
\shortstack{REF: \\ 0.55 \\ (-0.09)} & 
\shortstack{\textbf{\ul{REF:}} \\ \textbf{\ul{58.1\%}} \\ (-1.7)} & 
\shortstack{REF: \\ 0.57 \\ (+0.07)} & 
\shortstack{\textbf{\ul{REF:}} \\ \textbf{\ul{58.1\%}} \\ (-1.7\%)} & 
\shortstack{\textbf{\ul{REF:}} \\ \textbf{\ul{0.66}} \\ (+0.03)} \\ \hline

\multirow{2}{*}{Multi Label} 
& P1 & \shortstack{71.0\% \\ (+3\%)} & \shortstack{0.75 \\ (-0.06)} & 
\shortstack{68.0\% \\ (0\%)} & \shortstack{\textbf{\ul{0.85}} \\ (+0.03)} & 
\shortstack{\textbf{\ul{73.0\%}} \\ (+8\%)} & \shortstack{0.77 \\ (0)} & 
\shortstack{69.0\% \\ (-1\%)} & \shortstack{0.68 \\ (-0.03)} & 
\shortstack{70.0\% \\ (-1\%)} & \shortstack{0.79 \\ (+0.08)} & 
\shortstack{72.0\% \\ (+2\%)} & \shortstack{0.75 \\ (-0.06)} & 
\shortstack{71.3\% \\ (+2.3\%)} & \shortstack{0.74 \\ (+0.03)} \\ \cline{2-16}

& P2 & \shortstack{70.0\% \\ (-2\%)} & \shortstack{0.75 \\ (+0.02)} & 
\shortstack{66.0\% \\ (-7\%)} & \shortstack{\textbf{\ul{0.86}} \\ (+0.09)} & 
\shortstack{\textbf{\ul{72.0\%}} \\ (+7\%)} & \shortstack{0.83 \\ (+0.14)} & 
\shortstack{68.0\% \\ (+1\%)} & \shortstack{0.69 \\ (-0.03)} & 
\shortstack{71.0\% \\ (0\%)} & \shortstack{0.81 \\ (+0.03)} & 
\shortstack{70.0\% \\ (+1\%)} & \shortstack{0.75 \\ (-0.05)} & 
\shortstack{69.8\% \\ (+2.4\%)} & \shortstack{0.76 \\ (+0.02)} \\ \hline

\end{tabular}%
}
\end{table}
\end{landscape}

\begin{table}[htb]
\caption{F1 score for binary depression classification against other research results on E-DAIC and DAIC-WOZ development set.}
\label{tab:d-papers}
\resizebox{\columnwidth}{!}{%
\begin{tabular}{l|cc|c|}
\hline
\multicolumn{1}{|l|}{Reference}                                    & E-DAIC & DAIC-WOZ & Methods   \\ \hline
\multicolumn{1}{|l|}{Deepseek R1 (prompt 1) (T) (Ours)}    & 0.65 & 0.69 & ZS inference using the raw interview transcriptions \\ \hline
\multicolumn{1}{|l|}{Gemini 1.5 Flash (A) (prompt 3) (Ours)}  & 0.56 & 0.77 & ZS inference using the raw audio interviews \\ \hline
\multicolumn{1}{|l|}{Gemini 2.5 Pro (A+T) (prompt 3) (Ours)}  & 0.60 & 0.71 & ZS inference using both raw audio and transcription \\ \hline
\multicolumn{1}{|l|} {(Villatoro-Tello et al.)\cite{VillatoroTello2021ApproximatingTM}} & 0.59 & 0.56 & \begin{tabular}[c]{@{}c@{}} Fine-tuning BERT on a task-specific dataset derived \\ from the DAIC-WOZ interviews to predict \\depression from linguistic cues \end{tabular} \\ \hline
\multicolumn{1}{|l|}{BERT (Senn et al.)\cite{9871120}}                      & -    & 0.60 &\begin{tabular}[c]{@{}c@{}} Compared three individual BERT models\\ (BERT, RoBERTa, DistilBERT) on transcriptions and \\four ensembles with varying architectures.\end{tabular}\\ \hline
\multicolumn{1}{|l|}{BERT (Danner et al.)\cite{10354236}}                   & -    & 0.64 & 
\multirow{3}{*}{\begin{tabular}[c]{@{}c@{}}They trained and fine-tuned BERT for on the transcriptions. \\ They performed ZS inference  with GPT-3.5 and chatGPT-4\end{tabular}} \\ \cline{1-3}
\multicolumn{1}{|l|}{GPT-3.5 (Danner et al.)\cite{10354236}}                  & -    & 0.78 & \\ \cline{1-3}
\multicolumn{1}{|l|}{ChatGPT-4 (Danner et al.)\cite{10354236}}               & -    & 0.61 & \\ \hline
\multicolumn{1}{|l|}{GPT-4 (Hadzic et al.)\cite{doi:10.1080/18824889.2024.2342624}}                         & -    & 0.71 & \begin{tabular}[c]{@{}c@{}}They conducted ZS inference \\ with GPT-4 on the transcriptions\end{tabular}\\ \hline
\multicolumn{1}{|l|}{GPT-3.5-turbo P2+SMMR (Guo et al.)\cite{guo2024soullmateapplicationenhancingdiverse}}      & -    & 0.76 & \multirow{2}{*}{\begin{tabular}[c]{@{}c@{}}GPT-3.5-turbo and GPT-4-turbo were evaluated  using\\ ZS with a method called Stacked Multi-Model Reasoning \end{tabular}} \\ \cline{1-3}
\multicolumn{1}{|l|}{GPT-4-turbo P2+SMMR (Guo et al.)\cite{guo2024soullmateapplicationenhancingdiverse}}        & -    & 0.79 & \\ \hline
\end{tabular}
}
\end{table}

\begin{table}[h]
\centering
\caption{Binary PTSD classification against other research results on E-DAIC development and test set.}
\label{tab:ptsd-c}
\begin{tabular}{l|cc|c|}
\hline
\multicolumn{1}{|l|}{Reference}                 & BA   & F1   & Methods \\ \hline
\multicolumn{1}{|l|}{GPT-4o mini (prompt 2) (Ours)} & 76\%                                       & 0.68 & ZS inference using the raw interview transcriptions                                   \\ \hline
\multicolumn{1}{|l|}{Flores et al. \cite{10386191}}        & 70\% & 0.69 & \begin{tabular}[c]{@{}c@{}} Utilized bidirectional GRU model with self-attention. \\ Designed for multi-task learning, using temporal facial features. \end{tabular} \\ \hline
\multicolumn{1}{|l|}{Galatzer-Levy et al. \cite{galatzerlevy2023capabilitylargelanguagemodels}} & 74\% & 0.64 & \begin{tabular}[c]{@{}c@{}}Med-PaLM 2 utilized zero-shot inference, processing clinical \\ interview transcriptions to evaluate psychiatric conditions.\end{tabular} \\ \hline
\multicolumn{1}{|l|}{Wu et al. (CALLM) \cite{10614318}}    & 77\% & 0.70 &\begin{tabular}[c]{@{}c@{}}Fine-tuned a pre-trained DistilBERT model utilizing the expanded \\ training dataset  provided by their augmentation process.\end{tabular} \\ \hline

\end{tabular}
\end{table}

\newpage

\subsection{Limitations}
\textbf{Few Shot Learning:}
In our few-shot experiments targeting the audio modality, we opted not to use multiple audio samples for evaluation due to observed inconsistencies in the model's processing capabilities. Initially, the model demonstrated substantial variability when tasked with handling three audio files simultaneously within a single prompt, often failing to transcribe or comprehend the full content accurately. This inconsistency prompted us to exclude direct audio samples from few-shot testing. To further explore these challenges, we investigated the model's ability to summarize transcribed texts derived from audio inputs, which revealed further issues with accuracy and consistency. These findings underline the need for enhanced model refinement to ensure reliable handling and understanding of complex audio data in future implementations.

\textbf{{Comparative Evaluation:}}
Additionally, when comparing our results with existing studies, to our knowledge, we have not found any work concerning PTSD severity and multi-class classification tasks on the E-DAIC dataset, which are largely unexplored. For depression severity classification, existing studies often employ different labeling systems, such as using all numbers within the range (e.g., 0-24) as labels, which differs from our methodology. This diversity in approaches makes direct comparisons challenging.

Furthermore, it’s important to highlight that some relevant studies were not included in our comparative analysis. This exclusion is due to differences in datasets, training methods, models, or the metrics reported, which were not directly comparable to our Balanced Accuracy (BA) and F1 scores. 

\textbf{{LLM Fine-tuning:}} Fine-tuning is a process in which a pretrained model is further trained on task-specific data to improve its performance. Although this approach has the potential to specialize models for specific domains, it is not without limitations. In the following, we detail the challenges encountered during the fine-tuning process for the binary depression detection task.

\begin{enumerate} \item The E-DAIC dataset used for fine-tuning is both small and imbalanced, with 86 depression samples and 189 Non-depression samples. This imbalance makes it challenging for the model to learn effectively. With limited "depression" examples, the model struggles to generalize, ultimately failing to surpass the performance achieved by zero-shot (ZS) inference on unprocessed data, where no task-specific fine-tuning was conducted.

\item Large language models (LLMs) do not natively support fine-tuning with audio input data. Since detecting depression often relies heavily on acoustic cues, the inability to fine-tune using audio severely limits the model’s multimodal capabilities. Consequently, comparisons between audio and text modalities become inherently skewed, as the model can be specialized for text through fine-tuning but must remain at a less adapted, effectively zero-shot state for audio data.

\item Fine-tuning LLMs requires significant computational resources, expertise in hyperparameter tuning, and extensive trial-and-error. Given the constrained dataset and the nuanced nature of depression detection, these overheads do not guarantee performance improvements, making fine-tuning both resource-intensive and, in this case, yield-limited.
\end{enumerate}

\section{Conclusion}



This study has systematically evaluated the application of Large Language Models (LLMs) in mental health diagnostics, focusing on depression and PTSD using the E-DAIC dataset. We conducted experiments across a range of tasks to evaluate the comparative effectiveness of text and audio modalities and to investigate whether a multimodal approach could enhance diagnostic accuracy. Our research introduced custom metrics, the Modal Superiority Score (MSS) and the Disagreement Resolution Score (DRS), specifically designed to measure how the integration of modalities impacts model performance. The analysis consistently demonstrated that text modality excelled in most tasks, outperforming audio modality. However, the integration of text and audio modalities often led to improved outcomes, suggesting that combining these modalities could leverage the complementary strengths of each. The MSS and DRS metrics provided insights into the extent of these improvements, highlighting scenarios where multimodal approaches showed promise over single modalities.

While these findings underscore the potential of multimodal approaches for enhancing the performance of LLMs, it is essential to emphasize that this study is purely computational. The clinical utility of these methods remains untested and must be validated through rigorous controlled clinical studies to determine their effectiveness and reliability in real-world settings. Moreover, these results are derived from a single dataset (E-DAIC), which is relatively small and may not generalize to broader, more diverse populations or clinical environments. Variability in language, speech patterns, and diagnostic criteria across different datasets or real-world conditions could significantly impact model performance.

\section*{Acknowledgement}
We recognize and express our gratitude to the authors of \cite{ringeval2019avec2019workshopchallenge} and \cite{Gratch2014TheDA} for providing the Extended Distress Analysis Interview Corpus(E-DAIC) dataset and the Distress Analysis Interview Corpus(DAIC-WOZ) dataset

\end{document}